\documentclass[runningheads]{llncs}
\usepackage{graphicx}
\usepackage{amsmath}
\usepackage{amssymb}
\usepackage{mathtools}
\usepackage{lipsum} 
\usepackage{siunitx}
\usepackage{upgreek}
\usepackage[noend]{algpseudocode}
\usepackage{algorithm}
\usepackage{multirow}
\usepackage{anyfontsize}
\usepackage{diagbox}
\usepackage{url}
\usepackage{tabularx,booktabs}
\usepackage{tabto}
\usepackage{breqn}
\usepackage{bbm}
\usepackage{enumitem}
\usepackage{balance}
\usepackage{upgreek}
\usepackage{mathrsfs}
\usepackage{multirow, booktabs}
\usepackage{longtable}
\usepackage{lscape}
\usepackage{makecell}
\usepackage{hvfloat}
\usepackage{hyperref}

\usepackage{chngcntr}
\counterwithin{figure}{section}
\counterwithin{table}{section}

\begin{document}
%
\title{Univariate Skeleton Prediction in Multivariate Systems Using Transformers\thanks{Paper accepted at ECML PKDD 2024}}
%
%
\author{Giorgio Morales \and
John Sheppard} 
%
\authorrunning{G. Morales and J. Sheppard}
%
\institute{Gianforte School of Computing, Montana State University, USA
\email{giorgiomorales@ieee.org, john.sheppard@montana.edu} 
}
\maketitle              
%
\begin{abstract}
Symbolic regression (SR) methods attempt to learn mathematical expressions that approximate the behavior of an observed system.
However, when dealing with multivariate systems, they often fail to identify the functional form that explains the relationship between each variable and the system's response. 
To begin to address this, we propose an explainable neural SR method that generates univariate symbolic skeletons that aim to explain how each variable influences the system's response.
By analyzing multiple sets of data generated artificially, where one input variable varies while others are fixed, relationships are modeled separately for each input variable. 
The response of such artificial data sets is estimated using a regression neural network (NN).
Finally, the multiple sets of input--response pairs are processed by a pre-trained Multi-Set Transformer that solves a problem we termed Multi-Set Skeleton Prediction and outputs a univariate symbolic skeleton.
Thus, such skeletons represent explanations of the function approximated by the regression NN.
Experimental results demonstrate that this method learns skeleton expressions matching the underlying functions and outperforms two GP-based and two neural SR methods.
\end{abstract}
\section{Introduction}

Symbolic regression (SR) 
aims to identify mathematical equations or symbolic expressions that capture the underlying relationships and dynamics of the studied phenomena~\cite{contemporarySR}. 
One of the main advantages of the expressions learned by an SR model is that they are interpretable by humans and
allow for the identification of cause-effect relationships between the inputs and outputs of a system~\cite{SRinterpretable}. 
These techniques not only capture the behavior of empirical data into analytical equations but also 
reduce the computational complexity during the inference phase, and have more powerful extrapolation ability than black-box models~\cite{MartiusLampert2017:EQL}. 

A limitation of existing SR approaches lies in their primary focus on minimizing prediction errors rather than distilling the underlying equations that govern the system dynamics~\cite{SRthatscales}. 
Consequently, the generated equations may exhibit high complexity, effectively approximating the observed data but failing to correspond to the underlying equations~\cite{contemporarySR}. 
This limitation poses challenges when performing out-of-sample inference, where generalization may be ineffective.

SR methods are often based on population-based algorithms, especially genetic programming (GP)~\cite{SRbenchmark}.
Nevertheless, a notable drawback of GP-based SR methods is that they suffer from slow computation. 
Their computational inefficiency is related to the inherent complexity of the search space and the expensive iterative calls to numerical optimization routines after each generation~\cite{SRbenchmark}.
In addition, these methods do not consider past experiences, as they require learning each problem from scratch. 
As such, the obtained models do not benefit from additional data or insights from different equations, hindering their capacity for improvement and limiting their generalization capabilities~\cite{SRthatscales,SRtransformer}.

In response to these limitations, recent advancements have seen the emergence of neural SR methods as a promising alternative. 
These methods may utilize general-purpose pre-trained models that generate symbolic expressions through a straightforward forward pass and, possibly, a single call to a numerical optimization routine~\cite{SRthatscales,SRtransformer,SymbolicGPT}. 
Thus, neural SR methods offer a substantial time speedup compared to GP-based approaches.
However, there remains a gap in prediction accuracy between the two paradigms~\cite{SRtransformer}.

Although recent neural SR methods may achieve low prediction errors when dealing with multivariate systems, they struggle to represent the dependency of the system’s response and each independent variable correctly. 
As a first step in addressing this, our proposed method aims to generate univariate symbolic skeletons that estimate functional relationships for each variable with respect to the response.
A symbolic skeleton expression is an abstract representation of a mathematical expression that captures its structural form without setting specific numerical values.
In other words, we propose to generate univariate skeletons that constitute mathematical ``explanations'' describing the interaction between each independent variable and the system’s response.

Given a multivariate regression problem that can be expressed in terms of a mathematical equation, our method identifies univariate symbolic skeleton expressions for each variable.
The skeleton prediction process begins by training a regression model, such as a neural network (NN), that approximates the system's function.
This regression model is used to estimate the response of multiple randomly generated sets of points where only the variable of interest is allowed to vary while the remaining variables are held fixed. 
We then define a novel Multi-Set Transformer model, which is pre-trained on a large dataset of synthetic symbolic expressions, that receives as inputs the sets generated by the NN and is used to find the univariate skeletons. 


\subsubsection{Contributions} We hypothesize that our method will generate univariate skeletons that are more similar to those corresponding to the underlying equations in comparison to other SR methods.
The generated skeletons represent explanations of how each variable is related to the system's response.
Thus, from an explainability standpoint, producing more faithful univariate skeletons means that we can offer better explanations about the system's behavior.
In addition, the generated skeletons may be taken as building blocks to be used to estimate the overall function of the system. 
Our main contributions are summarized as: 

\begin{itemize}
    \item We introduce a method that 
    learns univariate symbolic skeletons that explain the functional form between each independent variable and the system's response using regression NNs and pre-trained transformer models. 
    \item We introduce a new SR problem called Multi-Set symbolic skeleton prediction (MSSP). It receives multiple sets of input--response pairs, where all sets correspond to the same functional form but use different equation constants, and outputs a common skeleton expression.  
    \item We present a novel transformer model called ``Multi-Set Transformer" to solve the MSSP problem. The model is pre-trained on a large dataset of synthetic symbolic expressions.  

\end{itemize}

\section{Related Work}

SR is commonly tackled using GP-based methods.
Such methods evolve a population of tree-like individuals using operations like selection, crossover, and mutation to improve their fitness over multiple generations.
Each individual represents a symbolic expression that maps the inputs and the output, and its fitness function determines how well it fits the data set being modeled.
Variations of this approach attempt to design improved operators and fitness functions to reduce the complexity of the search
~\cite{TaylorGP,SRbenchmark,eureqa}.

Two significant challenges of GP for SR are code growth (i.e., bloat) and the huge search space~\cite{statisticalGP}. 
Code growth leads to computationally expensive evolution of large programs and hinders their generalization ability~\cite{bloat1}. 
The huge search space is attributed to the variability in program lengths permitted during the evolutionary process.
This allows for the generation of multiple solution trees representing mathematically equivalent functions~\cite{searchspace}.
However, GP tends to generate a greater number of large solution trees compared to smaller ones~\cite{nature}.
The high complexity of large solutions entails poor generalization performance.  
In most GP-based methods, the program fitness is determined by its overall output and not by the intermediate outputs of its subexpressions. 
Thus, a program's subexpressions are optimized indirectly.
Arnaldo \textit{et al.}~\cite{MRGP} pointed out that the indirect optimization approach may result in suboptimal fitness optimization because it does not focus on evolving and learning suitable building blocks.

Furthermore, 
Martius and Lampert~\cite{MartiusLampert2017:EQL}, Sahoo \textit{et al.}~\cite{sahoo}, and Werner \textit{et al.}~\cite{iEQL} proposed the use of NN architectures whose units represent conventional mathematical operators.
The objective is to reduce the prediction error and gradually prune irrelevant parts of the network until a simple equation can be extracted from the network.
However, similar to the case of GP-based methods, the learned expressions tend to be large and, as a consequence, difficult to explain.
Another limitation is that each problem is learned from scratch. 

As a recent alternative, a few methods based on large language models have been proposed.
Biggio \textit{et al.}~\cite{SRthatscales} introduced the use of pre-trained transformer models for SR.
A dataset of multivariate equations is generated to pre-train a model based on a Set Transformer~\cite{settransformer}, and this transformer acts as a general-purpose model to predict the symbolic skeleton from a corresponding set of input-output pairs.  
The skeleton's constants are then fit using a non-convex optimizer, such as the BFGS algorithm~\cite{bfgs}. 

Kamienny \textit{et al.}~\cite{SRtransformer} pointed out that the loss function minimized by the BFGS algorithm can be highly non-convex and the correct constants of the skeletons are not guaranteed to be found.
They avoided performing skeleton prediction as an intermediary step and proposed an end-to-end (E2E) transformer model that estimates the full mathematical expression directly.
The learned constants can then be refined using a non-convex optimizer.
Experiments showed that E2E performs better than previous NN-based methods.
Since previous methods have shown scalability issues when dealing with equations having many variables, Chu \textit{et al.}\cite{SRcontrolvariables} proposed a method that decomposes multi-variable SR into a sequence of single-variable SR problems, combined in a bottom-up manner. 
The process involves learning a data generator using NNs from observed data, generating variable-specific samples with controlled input variables, applying single-variable SR, and iteratively adding variables until completion.

\section{Multivariate Skeleton Prediction}

Consider a system whose response $y \in \mathbb{R}$  is sensitive to variables $\textbf{x}=\{ x_1, \dots, x_t \}$ ($\textbf{x} \in \mathbb{R}^t$).
The underlying function that maps the feature value space and the response value space is denoted as $f:\mathcal{X}^t \rightarrow \mathcal{Y}$, such that $y = f(\mathbf{x}) = f(x_1, \dots, x_t)$. 
In addition, let $\kappa(\cdot)$ represent a skeleton function that replaces the numerical constants of a given symbolic expression by placeholders $c_i$; e.g., $\kappa (3x^2 +e^{2x} -4) = c_1\,x^2 + e^{c_2\, x} + c_3$.
We assume $f$ can be expressed as a mathematical expression with unary (e.g., $\texttt{sin}$, $\texttt{cos}$, and $\texttt{log}$) and binary operators (e.g. $+$, $-$, $*$, and $/$).
We aim to obtain the univariate skeletons $\hat{\mathbf{e}}(x_1),\dots, \hat{\mathbf{e}}(x_t)$ that describe the functional form between each variable and the system's response.
In the following, we lay out the steps proposed in this method. 

\subsection{Neural Network Training}  \label{sec:NN}

The underlying function $f$ can be approximated based on observed data using any regression model.
Let $\textbf{X}= \{ \textbf{x}_1, \dots , \textbf{x}_{N_R} \}$ be a data set with $N_R$ samples, where each sample is denoted as $\textbf{x}_j = \{ x_{j,1}, \dots, x_{j,t}\}$, and $\textbf{y}= \{ y_1, \dots , y_{N_R} \}$ is the set of corresponding target observations.
A NN regression model, whose compute function is denoted as $\hat{f}(\cdot; \boldsymbol{\theta}_{NN})$ ($\boldsymbol{\theta}_{NN}$ represents the weights of the network), is constructed to capture the association between $\textbf{X}$ and $\textbf{y}$.
Thus, a target estimate for a given input $\textbf{x}_j$ is computed as $\hat{y}_j = \hat{f}(\textbf{x}_j, \boldsymbol{\theta}_{NN})$ or, simply, $\hat{y}_j = \hat{f}(\textbf{x}_j)$.
The parameters $\boldsymbol{\theta}_{NN}$ of the network $\hat{y}$ are obtained by minimizing the mean squared error of the predictions; that is, 
$\boldsymbol{\theta}_{NN}^* = \; {\text{argmin}_{\boldsymbol{\theta}_{NN}}} \ \frac{1}{{N_R}} \sum_{j=1}^{N_R} (\hat{y}_{j} - y_{j})^2$. 
Note that we selected an NN to generate the function $\hat{f}$ due to its ease of training and high accuracy; however, other regression methods could be applied.

\subsection{Multi-Set Symbolic Skeleton Prediction}

\begin{figure}[t]
    \centering
    \includegraphics[width=0.75\columnwidth]{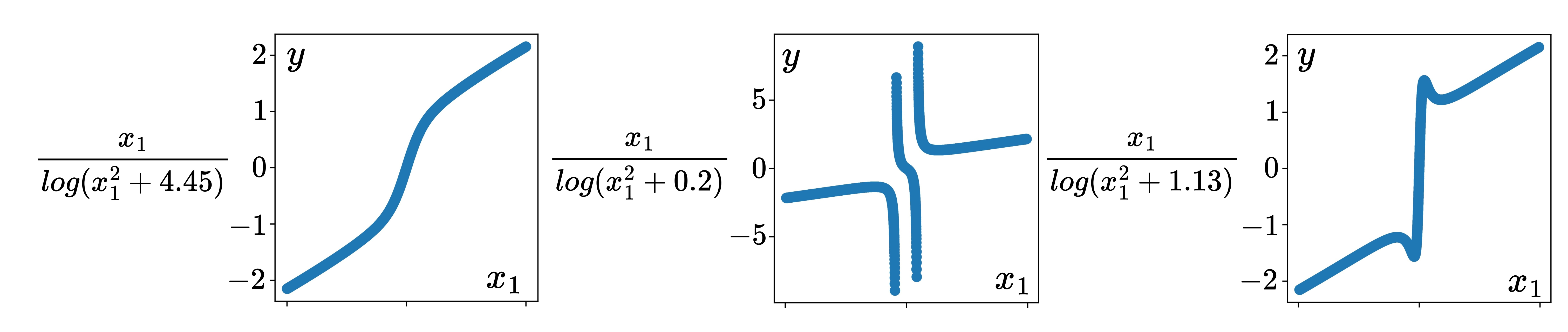}
    \caption{$x_1$ vs. $y$ curves when $x_2=4.45$, $0.2$, and $1.13$.}
    \label{fig:log_curve}
\end{figure}

We tackle the SR problem by decomposing it into single-variable sub-problems. 
To do so, we deviate from the Symbolic Skeleton Prediction (SSP) problem that has been explored previously by existing research~\cite{SRthatscales,petersen2021deep,SymbolicGPT}.
We explain the rationale behind this decision through the use of an example:
Consider the function $y = {x_1}/{\log (x_1^2 + x_2)}$. 
We analyze the relationship between $x_1$ and $y$. 
Similar to the work presented by Chu \textit{et al.}~\cite{SRcontrolvariables}, the variables that are not being analyzed are held fixed.
Fig.~\ref{fig:log_curve} depicts that the function's behavior differs when using three different $x_2$ values. 
As such, conventional SSP solvers may produce expressions with different functional forms depending on the fixed values of the variables not currently under analysis.

What is more, the fixed values of the remaining variables could project the function into a space where the functional form is not easily identifiable, which could be worsened due to the limited range of values that the analyzed variable can take. 
Therefore,  the SSP problem would benefit from the injection of additional context data.
When analyzing the variable $x_v$, we could use multiple sets of input--response pairs, each of which is constructed using different fixed values for the variables $\mathbf{x}\setminus \{x_v\}$.
The key idea is to process the information from these sets simultaneously to produce a skeleton that is common to all input sets.
We refer to this problem as Multi-Set Symbolic Skeleton Prediction (MSSP).

More formally, assume we are given a data set of $N_R$ input--response pairs $( \mathbf{X}, \mathbf{y})$ where $\mathbf{X} \in \mathbb{R}^{N_R \times t}$ and $\mathbf{y} \in \mathbb{R}^{N_R \times 1}$.
Suppose we analyze the relationship between the $v$-th variable of the system, $x_v$ (i.e., $v \in [1, \dots, t]$), and the response variable $y$. 
We construct a collection of $N_S$ sets, denoted as $\mathbf{D} = \{\mathbf{D}^{(1)}, \dots, \mathbf{D}^{(N_S)} \}$.
Each set $\mathbf{D}^{(s)}$ comprises $n$ input--response pairs such that $\mathbf{D}^{(s)} = ( \mathbf{X}_v^{(s)}, f(\mathbf{X}^{(s)}) ) = ( \mathbf{X}_v^{(s)}, \mathbf{y}^{(s)} )$,  where $\mathbf{X}^{(s)} \in \mathbb{R}^{n \times t}$, $\mathbf{y}^{(s)} \in \mathbb{R}^{n}$, and $\mathbf{X}_v^{(s)}$ denotes the $v$-th column of $\mathbf{X}^{(s)}$ (i.e., the data corresponding to the $x_v$ variable).  
$\mathbf{X}^{(s)}$ is constructed in such a way that each of the variables in $\mathbf{x}\setminus \{x_v\}$ are assigned random values and held fixed for all samples.

$\mathbf{X}^{(s)}$ can be constructed by selecting $n$ samples from $\mathbf{X}$ that meet the criterion that the variable $x_v$ is allowed to vary while the variables in $\mathbf{x}\setminus \{x_v\}$ are fixed. 
In this case, the set $\mathbf{y}^{(s)}$ is known as it is the corresponding set of observed response values of $\mathbf{X}^{(s)}$.
Conversely, if the set $\mathbf{X}$ is not large enough, $\mathbf{X}^{(s)}$ can be generated artificially using the regression model, and the corresponding set of response values would be obtained as $\mathbf{y}^{(s)} = f(\mathbf{X}^{(s)})$. 
Since $f$ is not known \textit{a priori}, the black-box model $\hat{f}$ that has been trained to approximate $f$ (i.e., $\hat{f}(\mathbf{x}) \approx f(\mathbf{x})$) is used to estimate the response of $\mathbf{X}^{(s)}$ (cf. Section \ref{sec:NN}).

Since the variables in $\mathbf{x}\setminus \{x_v\}$ have been fixed to constant values to construct each set $\mathbf{D}^{(s)}$, the underlying function that explains the relationship between $\mathbf{X}_v^{(s)}$ and $\mathbf{y}^{(s)}$ could be expressed solely in terms of variable $x_v$.
As such, the underlying function of the $s$-th set is denoted by $f^{(s)}(x_v)$.
It is important to note that functions $f^{(1)}(x_v),\dots, f^{(N_S)}(x_v)$ have been derived from the same function $f(\mathbf{x})$ and only differ in their coefficient values due to the selection of different values for the variables in $\mathbf{x}\setminus \{x_v\}$ for each constructed set.
Thus, if we apply the skeleton function $\kappa(\cdot)$ to functions $f^{(1)}(x_v),\dots, f^{(N_S)}(x_v)$, they should produce the same target symbolic skeleton $\mathbf{e}(x_v)$, in which the constant values have been replaced by placeholders; i.e., $\mathbf{e}(x_v) = \kappa\left( f^{(1)}(x_v) \right) = \dots = \kappa\left( f^{(N_S)}(x_v) \right)$.

Then, the collection $\mathbf{D}$ is fed as an input to the MSSP problem. The objective is to generate an skeleton $\hat{\mathbf{e}}(x_v)$ that characterizes the functional form of all input sets
, and approximates the target skeleton $\mathbf{e}(x_v)$; i.e., $\hat{\mathbf{e}}(x_v) \approx \mathbf{e}(x_v)$.
For the sake of generality, we define the MSSP problem as follows:

\begin{definition}
The input of the Multi-Set Symbolic Skeleton Prediction (MSSP) problem consists of a collection of $N_S$ sets, denoted as $\mathbf{D} = \{\mathbf{D}^{(1)}, \dots, \mathbf{D}^{(N_S)} \}$. Each set $\mathbf{D}^{(s)}$ comprises $n$ input--response pairs such that $\mathbf{D}^{(s)} = ( \mathbf{X}_{v}^{(s)}, \mathbf{y}^{(s)})$,  where $\mathbf{X}_{v}^{(s)} \in \mathbb{R}^{n}$ and $\mathbf{y}^{(s)} \in \mathbb{R}^{n}$. 
The underlying function of the $s$-th set is denoted by $f^{(s)}$; i.e., $\mathbf{y}^{(s)} = f^{(s)} (\mathbf{X}_v^{(s)})$.
The underlying functions of all input sets are assumed to share a common unknown symbolic skeleton, denoted as $\mathbf{e}$.
Thus, the objective of the MSSP problem is to generate a symbolic skeleton $\hat{\mathbf{e}} \approx \mathbf{e}$ that characterizes the functional form of all input sets $\mathbf{D}$.
\label{def:MSSP}
\end{definition}

\subsubsection{Multi-Set Transformer }\label{Sect:multisettransformer}

Our method for solving the MSSP problem draws inspiration from the Set Transformer~\cite{settransformer}, which is an attention-based neural network that is derived from the transformer model~\cite{attention}. 
The Set Transformer is designed for set-input problems and, as such, it is capable of processing input sets of varying sizes and exhibits permutation invariance.
We propose a Multi-Set Transformer that presents modifications to address the limitations of the Set Transformer and adapt it to the specific requirements of the proposed research.
Details about the differences between the Set Transformer and our Multi-Set Transformer are given in the \textbf{Supplementary Materials}\footnote{\label{supplementary}Please visit \url{https://github.com/NISL-MSU/MultiSetSR} for the supplementary file}.

Let the $s$-th input set be denoted as $\mathbf{D}^{(s)} = ( \mathbf{X}_v^{(s)}, \mathbf{y}^{(s)} ) = \{ (x_{v, i}^{(s)}, y_i^{(s)} ) \}_{i=1}^n$.
The first step involves arranging $\mathbf{D}^{(s)}$ in a manner analogous to the input structure of the Set Transformer, which consists of a matrix where each row represents a $d_{in}$-dimensional element of the input set.
$\mathbf{S}^{(s)} \in \mathbb{R}^{n \times d_{in}}$ denotes the $s$-th input of our proposed Multi-Set Transformer such that its $i$-th row, $\mathbf{s}_i^{(s)}$, consists of the concatenation of the input value $x_{v, i}^{(s)}$ and its corresponding output $y_i^{(s)}$; i.e., $\mathbf{s}_i^{(s)} = [x_{v, i}^{(s)}, y_i^{(s)}]$.
Hence, $\mathbf{S}^{(s)}$ is defined as a matrix with $d_{in} = 2$ columns.

Our Multi-Set Transformer comprises two primary components: an encoder and a decoder.
The encoder maps the information of all input sets into a unique latent representation $\mathbf{Z}$.
To do so, an encoder stack $\phi$ transforms each input set $\mathbf{S}^{(s)}$ into a latent representation $\mathbf{z}^{(s)} \in \mathbb{R}^{d}$  (where $d$ is context vector length or the ``embedding size") individually.
Our encoder, denoted as $\Phi$, comprises the use of the encoder stack $\phi$ to generate $N_S$ individual encodings $\mathbf{z}^{(1)}, \dots, \mathbf{z}^{(N_S)}$, which are then aggregated into a unique latent representation: 
\[
\mathbf{Z} = \Phi ( \mathbf{S}^{(1)}, \dots, \mathbf{S}^{(N_S)}, \boldsymbol{\theta}_e ) = \rho ( \phi ( \mathbf{S}^{(1)}, \boldsymbol{\theta}_e  ) , \dots, \phi ( \mathbf{S}^{(N_S)}, \boldsymbol{\theta}_e  ) ),
\]
where $\rho(\cdot)$ is a pooling function, $\boldsymbol{\theta}_e$ represents the trainable weights of the encoder stack, and $\mathbf{z}^{(s)} = \phi \left( \mathbf{S}^{(N_S)}, \boldsymbol{\theta}_e  \right)$.
We define $\phi$ as a stack of $\ell$ induced set attention blocks (ISABs)~\cite{settransformer} to encode high-order interactions among the elements of an input set in a permutation-invariant way. 
ISAB is a multi-head attention layer with low complexity ($\mathcal{O}(mn)$) that acts as a universal approximator of permutation invariant functions.
Furthermore, 
we include a pooling by multi-head attention (PMA) layer~\cite{settransformer} in $\phi$ to aggregate the features extracted by the ISAB blocks, whose dimensionality is $n \times d$, into a single $d$-dimensional latent vector.
Finally, the function $\rho(\cdot)$ that is used to aggregate the latent representations $\mathbf{z}^{(s)}$ is implemented using an additional PMA layer. 

The decoder $\psi$ generates sequences conditioned on the representation $\mathbf{Z}$ generated by $\Phi$.
This objective is aligned with that of the transformer decoder~\cite{attention} and, thus, the same architecture is used for our model.
$\psi$ consists of a stack of $M$ identical blocks, each of which is composed of three main layers: a multi-head self-attention layer, an encoder--decoder attention layer, and a position-wise feedforward network.
Let $\hat{\mathbf{e}} = \{ \hat{e}_1, \dots, \hat{e}_{N_{out}} \}$ denote the output sequence produced by the model, which represents the skeleton as a sequence of indexed tokens in prefix notation.
Each token in this sequence is transformed into a numerical index according to a pre-defined vocabulary that contains all unique symbols. 
During inference, each element $\hat{e}_i$ is generated auto-regressively; that is, $\psi$ produces a probability distribution over the elements of the vocabulary: $
\sigma \left( \psi \left(\mathbf{Z}, \boldsymbol{\theta}_d | \hat{e}_1, \dots, \hat{e}_{i-1} \right) \right)= P\left( \hat{e}_i | \hat{e}_1, \dots, \hat{e}_{i-1}, \mathbf{Z} \right)$,
where  $\boldsymbol{\theta}_d$ represents the weights of $\psi$.
This distribution is obtained by applying a softmax function $\sigma(\cdot)$ to the decoder's output.
$\hat{e}_i$ is thus selected as the token with the highest probability.

\subsubsection{Multi-Set Transformer Training} \label{sec:training}

We train our model on a large dataset of artificially generated MSSP problems.
Let $\mathbf{D}^b = \left\{ \mathbf{D_1}, \dots, \mathbf{D}_{B} \right\}$ denote a training batch with $B$ samples, each of which represents a collection of $N_S$ input sets; i.e., $\mathbf{D}_j = \{ \mathbf{D}_j^{(1)}, \dots, \mathbf{D}_j^{(N_S)} \}$ ($j \in [1, \dots, B]$). 
In addition, $\mathbf{E}^b = \{ \mathbf{e}_1, \dots, \mathbf{e}_{B} \}$ is the corresponding set of target skeletons, each of which represents a sequence of variable length; i.e., $\mathbf{e}_j = \left\{ e_{j,1} \dots, e_{j,N_{j}} \right\}$ and $N_j = |\mathbf{e}_j|$.
The function computed by the model is denoted as $g(\cdot)$, and $\boldsymbol{\Theta}$ denotes its weights.
Note that $\boldsymbol{\Theta} = [\boldsymbol{\theta}_e, \boldsymbol{\theta}_d]$ contains the weights of the encoder and the decoder stacks.
Given an input set collection $\mathbf{D}_j$, $g(\mathbf{D}_j, \boldsymbol{\Theta})$ computes the estimated skeleton $\hat{\mathbf{e}}_j$ with length $N_{out, \,j}$.
$g(\cdot)$ is trained to generate estimated skeletons so that $\hat{\mathbf{e}}_j \approx \mathbf{e}_j$.

During training, the model is provided with past elements of the target skeleton sequence as inputs for generating subsequent tokens.
Hence, the normalized probability distribution produced by the decoder $\psi$ for the $i$-th element of the predicted skeleton sequence of the $j$-th sample would be expressed as $P\left( \hat{e}_{j,i} | e_{j,1}, \dots, e_{j, i-1}, \mathbf{Z}_j \right)$.
In cases where  $\hat{\mathbf{e}}_j$ and $\mathbf{e}_j$ differ in length, we use padding and masking to ensure that the loss is only calculated for valid tokens.
Thus, our optimization objective is defined as the cross-entropy loss between the padded target and predicted skeleton sequences, which is calculated using their corresponding probability distributions over the set of possible tokens:

 \[ 
    \mathcal{L} = -\frac{1}{B}\sum_{j=1}^B \sum_{i=1}^T \omega_{j,i}  P(e_{j, i}) \log P( \hat{e}_{j,i} | e_{j,1}, \dots, e_{j, i-1}, \mathbf{Z}_j ),
 \]
where $w_{j, i}$ is 0 for padding positions and 1 otherwise, and $T_j$ is the length of the $j$-th sequence.
Hence, the optimization problem is expressed as:  $ \boldsymbol{\Theta} = \text{argmin}_{\boldsymbol{\Theta}} \ \mathcal{L}.$


We generate a training dataset of expressions stored in prefix notation.
The generation process is explained in Section B of the \textbf{Supplementary Materials}\hyperref[supplementary]{\textsuperscript{1}}.
Our method differs from the generation method used by Biggio \textit{et al.}~\cite{SRthatscales} and Lample \textit{et al.}~\cite{Lample2020Deep}.
Their approach generates several expressions that contain binary operators exclusively and, thus, are limited to simple expressions such as $\hat{e}(x)=c_1x$.
Conversely, we use a method that builds the expression tree recursively in preorder, which allows us to enforce certain conditions and constraints effectively.
That is, we forbid certain combinations of operators and set a maximum limit on the nesting depth of unary operators within each other.


The training routine of the Multi-Set Transformer $g$ takes as inputs the set of pre-generated expressions, denoted as $\mathbf{Q}$.
The skeleton $\mathbf{e}_j$ and its associated data collection $\mathbf{D}_j$, corresponding to the $j$-th expression in $\mathbf{Q}$, are generated using a function called $\texttt{generateSets}(\mathbf{Q}[j], N_S, n)$ (Algorithm~\ref{alg:generateData}). 
The estimated skeletons $\hat{\mathbf{e}}_j$ are obtained processing the input $\mathbf{D}_j$ and target $\mathbf{e}_j$ through the network $g$ using a teacher forcing strategy.
$\mathcal{L}(\mathbf{E}^B, \hat{\mathbf{E}}^B)$ represents the loss function while {\tt update}($g, L$) encompasses the conventional backpropagation and stochastic gradient descent processes used to update the weights of model $g$.

Algorithm~\ref{alg:generateData} describes the function \texttt{generateSets}, which takes as input a pre-generated expression $\mathbf{ex}$.
Function $\texttt{getConstants}(\mathbf{ex})$ returns $\mathbf{c}$, the list of constant placeholders in $\mathbf{ex}$, and $n_c = |\mathbf{c}|$.
Function $\texttt{selectConstants}(\mathbf{ex}, \mathbf{c}, n_f)$ retrieves an expression with $n_f$ constant placeholders selected randomly ($2 \leq n_f \leq n_c$).
Then, we generate the underlying functions $f^{(s)}$ corresponding to each of the $N_S$ sets using the function $\texttt{sampleConstants} (\mathbf{ex})$, which samples the values of each constant in $\mathbf{ex}$ from a uniform distribution $\mathcal{U}(-10, 10)$ independently.
The $n$ input points $\mathbf{X}_v^{(s)}$ are then sampled using the function $\texttt{sampleSupport}(n)$ from a uniform distribution $\mathcal{U}(-x^{\text{limit}}, x^{\text{limit}})$, where $x^{\text{limit}} \sim \mathcal{U}(1, 10)$ (i.e., the domain range is randomly sampled each time).
The function $\texttt{avoidNaNs}(\mathbf{X}_v^{(s)}, f^{(s)})$ may modify the coefficients in $f^{(s)}$ or sample additional $\mathbf{X}_v^{(s)}$ values to avoid undefined results (see Section B of the \textbf{Supplementary Materials}\hyperref[supplementary]{\textsuperscript{1}}).

\begin{algorithm} [t]
\fontsize{8}{9}\selectfont
\begin{algorithmic}[1]
\Function{generateSets}{$\mathbf{ex}, N_S, n$}
    \State $\mathbf{c}, n_c \leftarrow \texttt{getConstants}(\mathbf{ex})$ 
    \State $\mathbf{ex} \leftarrow \texttt{selectConstants}(\mathbf{ex}, \mathbf{c}, n_f = \texttt{randInt}(2, n_c))$ 
    \State $\mathbf{D}, s, \mathbf{e} \leftarrow [\,], 1, \kappa(\texttt{sampleConstants}(\mathbf{ex}))$
    \While {$s \leq N_S$}
        \State $\mathbf{X}_v^{(s)} \leftarrow \texttt{sampleSupport}(n)$
        \State $\mathbf{X}_v^{(s)}, f^{(s)}, \_\_ \leftarrow \texttt{avoidNaNs}(\mathbf{X}_v^{(s)}, \texttt{sampleConstants}(\mathbf{ex}))$
        \If{ $s > 1$ and $\kappa(f^{(s)}) \neq \mathbf{e}$}  
            \State \textbf{continue}  \Comment{Verify that all sets correspond to the same skeleton}
        \EndIf
        \State $\mathbf{y}^{(s)} \leftarrow f^{(s)}(\mathbf{X}_v^{(s)})$
        \State $\mathbf{D}.\texttt{append}( (\mathbf{X}_v^{(s)},\mathbf{y}^{(s)}))$
        \State $s \leftarrow s + 1$
    \EndWhile
    \State \Return $\mathbf{D}, \mathbf{e}$
\EndFunction
\end{algorithmic}
\caption{Multi-Set Transformer Data Generation}
\label{alg:generateData}
\end{algorithm}

\subsubsection{Univariate Symbolic Skeleton Prediction} \label{sec:univariateskeleton}

Suppose we are currently analyzing the $v$-th variable, $x_v$.
As an MSSP, we generate $N_S$ artificial sets of points $\{\tilde{\mathbf{X}}^{(1)}, \dots, \tilde{\mathbf{X}}^{(N_S)}\}$ where the variable $x_v$ is allowed to vary while the variables $\mathbf{x}\setminus \lbrace x_v\rbrace$ are fixed to random values. 
Specifically, the $s$-th artificial set is denoted as $\tilde{\mathbf{X}}^{(s)} = \{ \tilde{\mathbf{x}}^{(s)}_1, \dots, \tilde{\mathbf{x}}^{(s)}_n\}$. 
The value of the $v$-th dimension of the $j$-th sample is obtained by sampling from the distribution $\mathcal{U} (x_v^{\min}, x_v^{\max})$ whose lower and upper bounds, $x_v^{\min}$ and $x_v^{\max}$, respectively, are calculated from the observed data.
The values assigned to the remaining dimensions are sampled independently using similar uniform distributions; however, the same value is shared across all samples  (i.e., $\tilde{\mathbf{x}}^{(s)}_{1,k} = \tilde{\mathbf{x}}^{(s)}_{2,k} = \dots = \tilde{\mathbf{x}}^{(s)}_{n,k}$, $\forall k\in[1, \dots, t]$ and $k \neq v$).

Our approach requires processing data sets where the variables not under analysis are held constant.
However, it is not always possible to find subsets of data that meet this condition for all variables in real-world datasets.
Even when such subsets exist, they may not be sufficiently large.
Thus, we generate sets of data $\tilde{\mathbf{X}}^{(s)}$ with the desired behavior and estimate their response using a regression model that has been trained on observed data.
Then, our method derives univariate skeletons based on multiple sets of input-estimated response pairs.
It is important to use a prediction model that learns a function $\hat{f}$ that is as close as possible to $f$ so that it accurately estimates how the real system would respond to the artificial inputs in $\tilde{\mathbf{X}}^{(s)}$.
As a consequence, our analysis can be regarded as an explainability method that generates univariate symbolic skeletons as explanations of the function approximated by the regression model.

The response of the samples in $\tilde{\mathbf{X}}^{(s)}$, $\tilde{\mathbf{y}}^{(s)}$, is estimated using the network $\hat{f}$ as $\tilde{\mathbf{y}}^{(s)} = \hat{f}(\tilde{\mathbf{X}}^{(s)})$.
In addition, $\tilde{\mathbf{D}}_v^{(s)} = ( \tilde{\mathbf{X}}_v^{(s)}, \tilde{\mathbf{y}}^{(s)} )$ denotes the set of $n$ input--response pairs and is used to analyze the relationship between the system's response and the $v$-th variable.
The collection of $N_S$ sets $\tilde{\mathbf{D}}_v = \{ \tilde{\mathbf{D}}_v^{(1)}, \dots, \tilde{\mathbf{D}}_v^{(N_S)} \}$ is then fed into the pre-trained Multi-Set Transformer $g$ so that the estimated skeleton obtained for variable $x_v$ is calculated as $\tilde{\mathbf{e}}(x_v) = g(\tilde{\mathbf{D}}_v, \boldsymbol{\Theta})$. 
This process is repeated for all the variables of the system to obtain their corresponding symbolic skeleton expressions with respect to the system's response.

\subsection{Performance Evaluation} \label{sec:performance}

Typically, SR methods are evaluated on benchmark datasets such as the Feynman SR Benchmark (FSRB)~\cite{feynman}.
Their performance is compared based on the mean squared error (MSE) achieved by the learned expressions on a subset of the available data. 
Note that a function that was learned by minimizing the prediction error with respect to the system's response does not necessarily correspond to the underlying functional form; e.g., $f(x)=\cos(-\frac{x}{10})^2$ and $\hat{f}(x)= -0.0093x^2 + 0.9983$ produce an MSE lower than $4\times 10^{-5}$ when $x \in [-5, 5]$. 
As such, conventional evaluation approaches may not align with our objectives. 

Instead of minimizing prediction error, our method produces a set of univariate skeletons aiming to describe the functional form between each variable and the system's response.
We argue that an expression with minimum prediction error can be produced as a consequence of identifying the correct functional form.
To the best of our knowledge, no previous work has addressed the problem of testing how well the learned expression's functional form (i.e., its skeleton) matches the system's underlying functional form.
Thus, we present a method to test the similarity between the underlying skeleton corresponding to the variable $x_v$, represented as $\textbf{e}(x_v) = \kappa(f(\textbf{x}), x_v)$, and the estimated skeleton $\hat{\mathbf{e}}(x_v)$.
The function $\kappa(\cdot, x_v)$ replaces the numerical constants of a given symbolic expression with the placeholder labeled with the suffix $c$.
In addition, it considers the remaining variables $\mathbf{x}\setminus x_v$ as constants as they are irrelevant when describing the functional form between $x_v$ and the system's response. For example, if $f(\mathbf{x}) = 3x_1^2 +\sqrt{x_2+1} / e^{2\, x_3}$, then  $\kappa(f(\textbf{x}), x_1)= c_1\,x_1^2 + c_2$.

We assign random numerical values to the coefficients of skeleton $\mathbf{e}(x_v)$ using the $\texttt{sampleConstants}$ routine (Algorithm~\ref{alg:generateData}) in order to obtain a function $f_{\text{target}}(x_v)$.
Let $f_{\text{est}}(x_v) = \texttt{setConstants}(\hat{\mathbf{e}}(x_v), \mathbf{c})$ denote the function obtained when replacing the $n_c$ constant placeholders in $\hat{\mathbf{e}}(x_v)$ with the numerical values in a given set $\mathbf{c}=[c_1, \dots, c_{n_c}]$.
If the functional form of $\hat{\mathbf{e}}(x_v)$ is mathematically equivalent to that of $\mathbf{e}(x_v)$, then there exists a set of values $\mathbf{c}$ so that the difference between $f_{\text{target}}(x_v)$ and $f_{\text{est}}(x_v)$ is zero for all values of $x_v$.
The optimal set $\mathbf{c}^*$ is found as: $\mathbf{c}^* = \text{argmin}_{\mathbf{c}} \sum_{x_v \in \mathbf{X}_v^{\text{test}}} |f_{\text{target}}(x_v) - f_{\text{est}}(x_v)|$, where $\mathbf{X}_v^{\text{test}}$ is a test set of $N_{\text{test}}$ elements whose elements are drawn from a distribution $\mathcal{U}(2\, x_v^{\min}, 2\, x_v^{\max})$.
Note that the domain of $\mathbf{X}_v^{\text{test}}$ is larger than that used for training (i.e., $[x_v^{\min}, x_v^{\max}]$).
If the estimated skeleton matches the system's underlying functional form, it should do so regardless of the variable domain.

The coefficient fitting problem is solved using a simple genetic algorithm (GA)~\cite{holland-genetic-algorithms-1992}.
The individuals of our GA are arrays of $n_c$ elements that represent potential $\mathbf{c}$ sets.
The population size is set to 500 and the optimization algorithm is stopped when the change of the objective function after 20 generations is less than $10^{-5}$.
We utilize tournament selection, binomial crossover, and generational replacement.
We selected this configuration as it demonstrated effective optimization results across all experiments utilized in this work.

If $\hat{\mathbf{e}}(x_v)$ and $\mathbf{e}(x_v)$ are not equivalent, the error $r = \sum_{x_v \in \mathbf{X}_v^{\text{test}}} |f_{\text{target}}(x_v) - \texttt{setConstants}(\hat{\mathbf{e}}(x_v), \mathbf{c}^*)|$ is greater than 0.
We use $r$ as a performance metric that indicates the closeness between $\hat{\mathbf{e}}(x_v)$ and $\mathbf{e}(x_v)$ given the sampled values of the constants of $\mathbf{e}(x_v)$.
Note that if $\hat{\mathbf{e}}(x_v)$ and $\mathbf{e}(x_v)$ are similar, the error $r$ should be low regardless of the sampled values of the constants of $\mathbf{e}(x_v)$.
Thus, for the sake of generality, we repeat this process 30 times; that is, we sample 30 different $f_{\text{target}}(x_v)$ functions and solve 30 optimization problems.
Finally, we report the mean and the standard deviation of the 30 resulting error metrics.

\section{Experimental Results}

A training dataset\footnote{The code and datasets are available at \url{https://github.com/NISL-MSU/MultiSetSR}} 
consisting of one million pre-generated expressions ($|\mathbf{Q}|=10^6$) has been created to train the Multi-Set Transformer. 
These expressions allow up to one nested operation and contain a maximum of five unary operators.
We also generated an independent validation set consisting of $10^5$ expressions.
For the model architecture, due to the high computational expense associated with training a single model, we used a one-factor-at-a-time approach to choose the following hyperparameters: the number of ISAB encoder blocks $\ell  = 3$, the number of decoder blocks $M = 5$, an embedding size $d = 512$, and the number of heads $h=8$.
In addition, we set the number of input sets to $N_S=10$ and the number of input-response pairs in each input set to $n=3000$. 
Future work will examine the impact of altering the values of $N_S$ and $n$ on the final performance.

\subsection{Experiments with Synthetic Data}

\begin{table}[t]
    \caption{Equations used for experiments}
    \label{tab:datasets}
    \centering
\resizebox{0.8\textwidth}{!}{%
\begin{tabular}{|c|c|c|c|}
\hline
\textbf{Eq.} & \textbf{Underlying equation}& \textbf{Reference} & \textbf{Domain range} \\ \hline
E1 & $ (3.0375 x_1 x_2 + 5.5 \sin (9/4 (x_1 - 2/3)(x_2 - 2/3)))/5$ & \cite{A1} & $[-5, 5]^2$\\ \hline
E2 & $5.5 + (1- x_1/4) ^ 2 + \sqrt{x_2 + 10} \sin( x_3/5) $ & --- & $[-10, 10]^2$ \\ \hline
E3 & $(1.5 e^{1.5  x_1} + 5 \cos(3 x_2)) / 10$ & \cite{A1} & $[-5, 5]^2$ \\ \hline
E4 & $((1- x_1)^2 + (1- x_3) ^2 + 100 (x_2 - x_1 ^ 2) ^ 2 + 100 (x_4 - x_3 ^ 2) ^ 2)/10000$ & Rosenbrock-4D & $[-5, 5]^4$ \\ \hline
E5 & $\sin(x_1 + x_2 x_3) + \exp{(1.2  x_4)}$ & --- & \begin{tabular}[c]{@{}c@{}}$x_1 \in [-10, 10],\, x_2 \in [-5, 5],$ \\ $\, x_3 \in [-5, 5], \, x_4 \in [-3, 3]$\end{tabular} \\ \hline
E6 & $\tanh(x_1 / 2) + |x_2|  \cos(x_3^2/5)$& --- & $[-10, 10]^3$ \\ \hline
E7 & $(1 - x_2^2) / (\sin(2 \pi \, x_1) + 1.5)$ & \cite{iEQL} & $[-5, 5]^2$ \\ \hline
E8 & $x_1^4 / (x_1^4 + 1) + x_2^4 / (x_2^4 + 1)$ & \cite{TRUJILLO201621} & $[-5, 5]^2$ \\ \hline
E9 & $\log(2 x_2 + 1) - \log(4 x_1 ^ 2 + 1)$ & \cite{TRUJILLO201621} & $[0, 5]^2$ \\ \hline
E10 & $\sin(x_1 \, e^{x_2})$ & \cite{metric} & $x_1 \in [-2, 2], x_2 \in [-4, 4]$ \\ \hline
E11 & $x_1 \, \log(x_2 ^ 4)$ & \cite{metric} & $[-5, 5]^2$ \\ \hline
E12 & $1 + x_1 \, \sin(1 / x_2)$ & \cite{metric} & $[-10, 10]^2$ \\ \hline
E13 & $\sqrt{x_1}\, \log(x_2 ^ 2)$& \cite{metric} & $x_1 \in [0, 20], x_2 \in [-5, 5]$ \\ \hline
\end{tabular}%
}
\end{table}

We assessed the performance of the skeletons generated by our Multi-Set Transformer using ten synthetic SR problems generated by equations inspired by previous works and equations proposed in this work, as reported in Table~\ref{tab:datasets}.
Note that previous works used narrow domain ranges for all variables (e.g., $[-1, 1]$) while we used extended ranges (e.g., $[-5, 5]$ and $[-10, 10]$) to increase the difficulty of the problems.
We adapted the benchmark equations proposed by Bertschinger \textit{et al.}~\cite{metric} (i.e., E10--E13) to a multivariate setting.

In all cases, the training datasets consisted of 10,000 points where each variable was sampled using a uniform distribution.
For the estimated response functions, $\hat{f}$, we trained feed-forward NNs with varying depths: three hidden layers for problem E2; five hidden layers for problems E1, E4, E5, and E7; and four hidden layers for the other cases.
Each layer consisted of 500 nodes with ReLU activation. 
We used 90\% of the samples for training and 10\% for validation.

We compared the skeletons produced by the Multi-Set Transformer (MST) to the ones extracted from the expressions generated by three other methods: two GP-based methods (PySR~\cite{pysr2} and TaylorGP~\cite{TaylorGP}) and two neural SR methods (NeSymReS~\cite{SRthatscales} and E2E~\cite{SRtransformer}).
For NeSymReS and E2E, we used the pre-trained models available online.
NeSymReS, E2E, and the Multi-Set Transformer were trained using vocabularies with the same unary and binary operators: $+, \times, /$, \texttt{abs}, \texttt{acos}, \texttt{asin}, \texttt{atan}, \texttt{cos}, \texttt{cosh}, \texttt{exp}, \texttt{log}, \texttt{pow2}, \texttt{pow3}, \texttt{pow4}, \texttt{pow5}, \texttt{sin}, \texttt{sinh}, \texttt{sqrt}, \texttt{tan}, \texttt{tanh}. 
Thus, PySR and TaylorGP were executed using the same set of operators.
Our experimentation with the GP-based methods involved a maximum of 10,000 iterations, though convergence was consistently achieved in fewer iterations across all cases.
Population sizes of 100, 200, 500, and 1000 were tested, with no discernible advantage observed beyond a size of 500. 

The compared methods produce multivariate expressions, from which the skeleton variable corresponding to variable $x_v$ is obtained using the skeleton function $\kappa(\cdot, x_v)$.
NeSymReS could not be executed on E4 and E5 because its model was trained using expressions limited to three variables.
Table~\ref{tab:results_skeletonsE2} shows the target and estimated skeletons corresponding to each variable for problem E2.
The skeletons obtained for the other problems are presented in Section C of the \textbf{Supplementary Materials}\hyperref[supplementary]{\textsuperscript{1}}.
We also evaluated skeleton performance using the 
method described in Section \ref{sec:performance}.
We set the size of the test sets to $N_{\text{test}}=3000$. 
Using $N_{\text{test}}>3000$ did not vary the obtained results.
Table~\ref{tab:skeleton_evaluation} reports the rounded mean and the standard deviation of the error metrics obtained after 30 repetitions of the proposed evaluation.  
The bold entries indicate the method that achieved the lowest mean error $r$ and that its difference w.r.t. the values obtained by the other methods is statistically significant according to Tukey’s honestly significant difference test performed at the 0.05 significance level.

\begin{table}[t]
    \caption{Comparison of skeleton prediction results for problem E2}
    \label{tab:results_skeletonsE2}
\centering
\Large
\resizebox{0.75\textwidth}{!}{%
\begin{tabular}{|c|c|c|c|}
\hline
\textbf{Method} & $x_1$ & $x_2$ & $x_3$  \\ \Xhline{4\arrayrulewidth}
\textbf{PySR} & $c_1 +\left|c_2 + |c_3 + x_1|\right|$ & $c_1$ & $c_1 + c_2\,x_3$\\ \hline
\textbf{TaylorGP} & $c_1 + c_2\,x_1$ & $c_1 + c_2\,x_2$  & $c_1 + c_2\,x_3$ \\ \hline
\textbf{NeSymReS} & $c_1 + c_2\,x_1$ & $c_1 + \exp(\exp(c_2\,x_2))$ & $c_1 + c_2 \, x_3$\\ \hline
\textbf{E2E} & $c_1 + c_2\, x_1 + c_3(c_4 + c_5\, x_1)^2$ & $c_1 + c_2\,(c_3 + c_4\, x_2)$ & $c_1 + c_2 \, x_3 + c_3 (c_4 + c_5 \cos(c_6 + c_7 \, x_3))$\\ \hline
\textbf{MST} & $c_1 + c_2(c_3 + c_4\, x_1)^2$ & $c_1\sqrt{c_2\,x_2 + c_3} + c_4$ & $c_1 + c_2 \sin(c_3\, x_3 + c_4)$\\ \hline
\textbf{Target} $\mathbf{e}(x)$ & $c_1 + (c_2 + c_3\,x_1)^2$ & $c_1\sqrt{x_2 + c_2} + c_3$ & $c_1 + c_2\sin(c_3\,x_3)$\\ \hline
\end{tabular}%
}
\end{table}

\begin{table}[t]
    \caption{Skeleton evaluation performance comparison}
    \label{tab:skeleton_evaluation}
    \centering
     \resizebox{0.65\textwidth}{!}{
    \begin{tabular}{|c|c|c|c|c|c|c|}
    \hline
    \textbf{Eq.}& \textbf{Var.} & \textbf{PySR}  &\textbf{TaylorGP}& \textbf{NeSymReS} & \textbf{E2E} & \textbf{MST}\\ \Xhline{4\arrayrulewidth}
    \multirow{2}{*}{\textbf{E1}} & $x_1$ &   $1.4 \pm 0.8$&$1.4 \pm 0.8$&  $0.9 \pm 0.7$&  $\mathbf{0.2 \pm 0.4}$&  $\mathbf{0.01 \pm 0.02}$\\ \cline{2-7} 
     & $x_2$ &   $1.5 \pm 0.9$&$1.5 \pm 0.9$&  $1.3 \pm 0.8$&  $1.5 \pm 0.9$&  $\mathbf{0 \pm 0}$
\\ \hline
    \multirow{3}{*}{\textbf{E2}} & $x_1$ &  $303.5 \pm 167.3$&$310.0 \pm 170.1$& $310.0 \pm 170.1$& $\mathbf{0 \pm 0}$& $\mathbf{0 \pm 0}$\\ \cline{2-7} 
     & $x_2$ &  $5.4 \pm 5.0$&$4.2 \pm 5.3$& $4.6 \pm 5.0$& $4.2 \pm 5.3$& $\mathbf{0.02 \pm 0.03}$\\ \cline{2-7} 
     & $x_3$ &   $1.7 \pm 1.0$&$1.7 \pm 1.0$&  $1.7 \pm 1.0$&  $\mathbf{0.01 \pm 0.02}$&  $\mathbf{0 \pm 0}$\\ \hline
    \multirow{2}{*}{\textbf{E3}} & $x_1$ &   $2\!\times\! 10^{12} \pm 5\!\times\! 10^{12}$&$939.4 \pm 1419.9$&  $1.9 \pm 1.2$&  $\mathbf{0.8 \pm 1.8}$&  $\mathbf{0.8 \pm 1.8}$\\ \cline{2-7} 
     & $x_2$ &   $1.3 \pm 1.0$&$1.3 \pm 1.0$&  $0.8 \pm 0.8$&  $\mathbf{0 \pm 0}$&  $\mathbf{0 \pm 0}$\\ \hline
    \multirow{4}{*}{\textbf{E4}} & $x_1$ &   $4576.2 \pm 2695.7$&$4581.5 \pm 2697.4$&  ---&  $\mathbf{2.3 \pm 3.6}$&  $\mathbf{1.1 \pm 0.7}$\\ \cline{2-7} 
     & $x_2$ &   $79.6 \pm 41.3$&$80.2 \pm 40.8$&  ---&  $\mathbf{0 \pm 0}$&  $\mathbf{0 \pm 0}$\\ \cline{2-7} 
     & $x_3$ &  $3995.5 \pm 2815.6$&$4304.6 \pm 2843.7$& \multicolumn{1}{c|}{---} & $\mathbf{2.0 \pm 4.0}$& $\mathbf{1.0 \pm 0.9}$\\ \cline{2-7} 
     & $x_4$ &  $74.5 \pm 48.0$&$75.5 \pm 47.0$& \multicolumn{1}{c|}{---} & $\mathbf{0 \pm 0}$& \multicolumn{1}{c|}{$\mathbf{0 \pm 0}$} \\ \hline
    \multirow{4}{*}{\textbf{E5}}& $x_1$ &   $0.6 \pm 0.05$&$0.6 \pm 0.05$&  ---&  $\mathbf{0 \pm 0}$&  $\mathbf{0 \pm 0}$\\
 \cline{2-7} 
& $x_2$ &  $1.5 \pm 1.0$&$1.5 \pm 1.0$& ---& $1.5 \pm 1.0$&$\mathbf{0 \pm 0}$\\      \cline{2-7} 
& $x_3$ &   $0.6 \pm 0.05$&$0.6 \pm 0.05$&  ---&  $0.6 \pm 0.05$&  
$\mathbf{0 \pm 0}$\\ \cline{2-7} 
& $x_4$&   $2.7 \pm 1.2$&$487.4 \pm 461.9$&  ---&  $\mathbf{0.6 \pm 0.8}$&  $\mathbf{0.6 \pm 0.8}$\\ \hline
    \multirow{3}{*}{\textbf{E6}}& $x_1$ &   $0.8 \pm 0.08$&$0.8 \pm 0.08$&  $0.3 \pm 0.01$&  $0.04 \pm 0$&  $\mathbf{0 \pm 0}$\\
 \cline{2-7} 
& $x_2$ &  $16.8 \pm 12.2$&$16.8 \pm 12.2$& $13.8 \pm 11.1$& $1.3 \pm 0.9$&$\mathbf{0 \pm 0}$\\      \cline{2-7} 
& $x_3$ &   $2.9 \pm 1.3$&$1.9 \pm 0.7$&  $1.6 \pm 0.9$&  $1.6 \pm 0.9$&  
$\mathbf{0 \pm 0}$\\ \hline
    \multirow{2}{*}{\textbf{E7}}& $x_1$ &   $29.5 \pm 1.0$&$1.8 \pm 2.0$&  $1.8 \pm 2.0$&  $1.1 \pm 1.3$&  $\mathbf{0 \pm 0}$\\
 \cline{2-7} 
& $x_2$ &  $63.6 \pm 43.8$&$1.6 \pm 1.0$
& $42.4 \pm 24.7$& $\mathbf{0 \pm 0}$& $\mathbf{0 \pm 0}$\\   \hline
    \multirow{2}{*}{\textbf{E8}}& $x_1$ &   $0.02 \pm 0.01$&$0.04 \pm 0.01$&  $0.8 \pm 1.2$&  $0.02 \pm 0.02$&  $\mathbf{0 \pm 0}$\\
 \cline{2-7} 
& $x_2$ &  $0.02 \pm 0.01$&  $0.04 \pm 0.01$& $0.9 \pm 1.3$& $0.02 \pm 0.01$& $\mathbf{0 \pm 0}$\\   \hline
    \multirow{2}{*}{\textbf{E9}}& $x_1$ &   $271.3 \pm 446.8$&$239.8 \pm 428.2$&  $375.1 \pm 485.5$&  $\mathbf{0 \pm 0}$&  $\mathbf{0 \pm 0}$\\
 \cline{2-7} 
& $x_2$ &  $\mathbf{0 \pm 0}$&$0.2 \pm 0.09$& $2.7 \pm 1.7$& $0.05 \pm 0.01$&$\mathbf{0 \pm 0}$\\   \hline
    \multirow{2}{*}{\textbf{E10}}& $x_1$ &   $\mathbf{0 \pm 0}$&$0.6 \pm 0.2$&  $\mathbf{0 \pm 0}$&  $\mathbf{0 \pm 0}$&  $\mathbf{0 \pm 0}$
\\
 \cline{2-7} 
& $x_2$ &  $\mathbf{0 \pm 0}$&$0.4 \pm 0.06$& $\mathbf{0 \pm 0}$& $\mathbf{0 \pm 0}$& $\mathbf{0 \pm 0}$ 
\\   \hline
    \multirow{2}{*}{\textbf{E11}}& $x_1$ &   $\mathbf{0 \pm 0}$&$\mathbf{0 \pm 0}$&  $\mathbf{0 \pm 0}$&  $\mathbf{0 \pm 0}$&  $\mathbf{0 \pm 0}$
\\
 \cline{2-7} 
& $x_2$ &  $\mathbf{0 \pm 0}$&$\mathbf{0 \pm 0}$& $\mathbf{0 \pm 0}$& $\mathbf{0 \pm 0}$&$\mathbf{0 \pm 0}$
\\   \hline
    \multirow{2}{*}{\textbf{E12}}& $x_1$ &   $21.8 \pm 13.1$&$\mathbf{0 \pm 0}$&  $\mathbf{0 \pm 0}$&  $\mathbf{0 \pm 0}$&  $\mathbf{0 \pm 0}$
\\
 \cline{2-7} 
& $x_2$ &  $\mathbf{0 \pm 0}$&$2.4 \pm 1.5$& $2.5 \pm 1.6$& $\mathbf{0 \pm 0}$&$\mathbf{0 \pm 0}$\\   \hline
    \multirow{2}{*}{\textbf{E13}}& $x_1$ &   $\mathbf{0 \pm 0}$&$0.8 \pm 0.5$
&  $0.7 \pm 0.6$&  $0.7 \pm 0.8$&  $\mathbf{0 \pm 0}$\\
 \cline{2-7} 
& $x_2$ &  $\mathbf{0 \pm 0}$&$\mathbf{0 \pm 0}$& $3.8 \pm 3.5$& $\mathbf{0 \pm 0}$&$\mathbf{0 \pm 0}$\\   \hline
    \end{tabular}
    }
\end{table}

\subsection{Analysis of Nitrogen Fertilizer Response Curves} \label{sec:Nresponse}

In the previous section, we performed experiments based on synthetic datasets.
Recall that the objective of this work is to demonstrate that our method can explain the relationship between each system's variable and the system's response.
To do so, datasets with known underlying functions are needed.
However, in this section, we show the applicability of our approach in a real-world application.

In precision agriculture, crop yield prediction is a critical task with a significant societal impact~\cite{hegedus}.
Let $y(i, j)$ represent the observed yield at a field site with coordinates $(i, j)$.
Furthermore, let $\mathbf{x}(i, j)$ represent a set of multiple covariate factors, such as the nitrogen (N) fertilizer rate applied, and other topographic and meteorological features.
It describes the state of the field at position $(i, j)$, and potentially its neighboring areas.
The underlying yield function of the field is denoted as $f(\cdot)$ and $y(i, j) = f(\mathbf{x}(i, j))$.
Thus, the early crop yield prediction problem consists of generating predicted yield values $\hat{y}(i, j)$ using a prediction model $\hat{f}(\cdot)$, such that $\hat{y}(i, j) = \hat{f}(\mathbf{x}(i, j))$ and $\hat{y}(i, j) \approx y(i, j)$.

In practice, $f$ is a complex multivariate system with unknown functional form.  
Nevertheless, tasks like N-rate optimization, which allows for profit maximization and environmental impact maximization~\cite{hegedus}, do not require estimating the full functional form of $f(\mathbf{x}(i, j))$.
Instead, N-rate optimization only analyzes the functional relationship between the N-rate variable and the predicted yield values.
This relationship is typically represented using N fertilizer-yield response (N-response) curves, which depict the estimated crop yield values for a specific field site in response to various admissible fertilizer rates.

The experiments presented in this section aim to estimate the functional form of N-response curves for a winter wheat dryland field called ``Field A", discussed in a previous work~\cite{MZ}.
Traditionally, N-response curves are fitted employing a single parametric function for the entire field~\cite{bullock94,CF}.
However, previous work~\cite{MZ} demonstrates that the N-response functional form varies within the field, which can be clustered into regions known as management zones (MZs) based on the fertilizer responsivity of the field sites (i.e., the shape of the approximated N-response curves).
For details about the dataset, the generation of approximated N-response curves, and the clustering method, please see~\cite{MZ}.

\begin{figure}[t]
    \centering
    \includegraphics[width=0.75\textwidth]{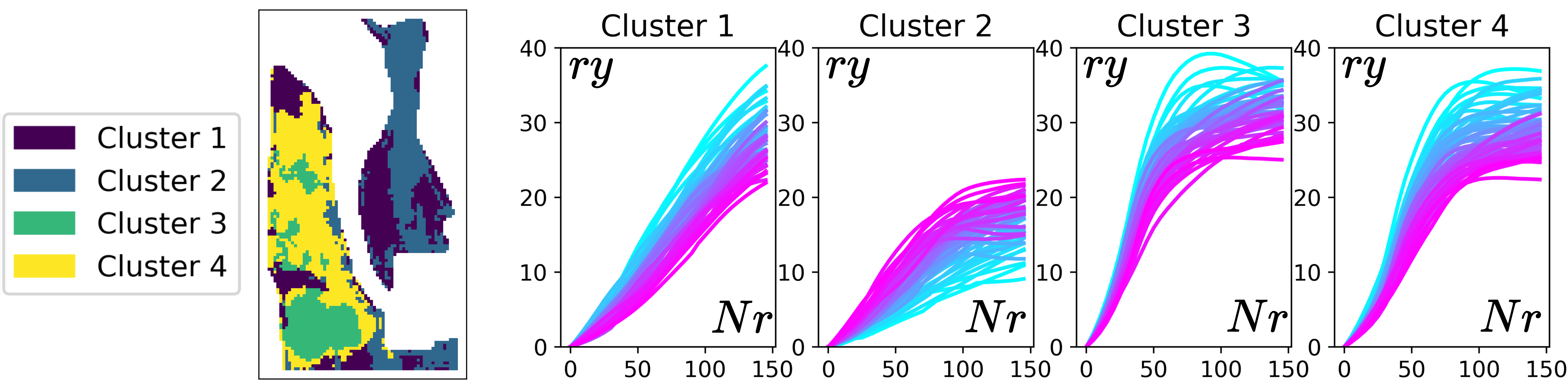}
    \caption{MZs for Field A and corresponding N-response curves~\cite{MZ}.}
    \label{fig:MZ}
\end{figure}

Fig.~\ref{fig:MZ} depicts Field A clustered into four MZs.
The figure also shows fifty example response curves selected randomly from each MZ.
Due to their shape similarity, we assume that all sites within an MZ share the same functional form.
Consider that an N-response curve is a set of input--response pairs (N-rate $Nr$ vs. relative yield $ry$).
Thus, the problem of estimating the functional form of the N-response curves of an MZ can be posed as an MSSP.
To do so, we select $N_S=10$ random field sites within an MZ and feed their corresponding N-response curves into the pre-trained MST model to obtain a univariate skeleton. 

\begin{table}[t]
    \caption{Comparison of skeleton prediction results for problem E2}
    \label{tab:results_Nresponse}
\centering
\resizebox{1\textwidth}{!}{%
\begin{tabular}{|c|c|c|c|c|c|c|c|}
\hline
{Method} & {MZ} & {Functional Form} & {$\bar{r}$} & {Method} & {MZ} & {Functional Form} & {$\bar{r}$} \\ \Xhline{3\arrayrulewidth}
Quadratic-plateau & \multirow{3}{*}{1} & $c_1 + c_2\,(\min(Nr, c_3) + c_3)^2$& 0.0680  & Quadratic-plateau & \multirow{3}{*}{3} &  $c_1 + c_2\,(\min(Nr, c_3) + c_3)^2$& 0.0879\\ \cline{1-1} \cline{3-5} \cline{7-8} 
Exponential &  & $c_1 (1-\exp(c_2 + c_3\, Nr)) + c_4$&  0.2303& Exponential &  &  $c_1 (1-\exp(c_2 + c_3\, Nr)) + c_4$& 0.1965\\ \cline{1-1} \cline{3-5} \cline{7-8} 
\textbf{MST} &  & $c_1 + c_2 ( Nr -\log(c_3 + c_4\, Nr))$& 0.0489 & \textbf{MST} &  &  $c_1 + c_2\,Nr + c_3\,cos(c_4 + c_5\,Nr)$&  0.0479\\ \Xhline{3\arrayrulewidth}
Quadratic-plateau & \multirow{3}{*}{2} & $c_1 + c_2\,(\min(Nr, c_3) + c_3)^2$&  0.0725& Quadratic-plateau & \multirow{3}{*}{4} &  $c_1 + c_2\,(\min(Nr, c_3) + c_3)^2$& 0.0615\\ \cline{1-1} \cline{3-5} \cline{7-8} 
Exponential &  & $c_1 (1-\exp(c_2 + c_3\, Nr)) + c_4$ &  0.1825& Exponential &  &  $c_1 (1-\exp(c_2 + c_3\, Nr)) + c_4$& 0.2249\\ \cline{1-1} \cline{3-5} \cline{7-8} 
\textbf{MST} &  & $c_1 + c_2\, \texttt{tanh}(c_3 + c_4\, Nr)$&  0.0495& \textbf{MST} &  &  $c_1 + c_2\,Nr + c_3\,cos(c_4 + c_5\,Nr)$&  0.0237\\ \hline
\end{tabular}%
}
\end{table}

Table~\ref{tab:results_Nresponse} shows the skeletons derived by the MST for each MZ.
The process was repeated multiple times for each MZ, with different random sites selected as inputs for the MSSP problem in each iteration, resulting in the generation of equivalent skeletons. 
Furthermore, we evaluated the suitability of the obtained skeletons for each MZ and compared them to two traditional N-response models: quadratic-plateau~\cite{bullock94} and exponential~\cite{CF}.
For each method and field site, we fit the skeleton's coefficient values to minimize the distance to the corresponding N-response curve, following a similar procedure as the one described in Sec.~\ref{sec:performance}.
We report the average error $\bar{r}$ obtained considering all sites within each MZ. 
Results indicate that different MZs may be modeled using different functional forms, unlike traditional approaches that assume one for the entire field.

\section{Discussion}

Our method involves training a feedforward NN, which is then used to solve an MSSP problem for each system's variable using a Multi-Set Transformer.
This process produces univariate skeletons that describe the functional relationship between each variable and the system's response.
After evaluation of our univariate skeleton prediction method across the tested problems, we observed that it generated skeletons that matched or were equivalent to the target skeleton for all variables across all problems.
For instance, for problem E11, the target skeleton for variable $x_2$ is given by $\mathbf{e}(x_2) = c_1\, \log(x_2^4)$, and our method produces the skeleton $\mathbf{\hat{e}}_{MST}(x_2) = c_1' + c_2'\, \log(c_3' \, x_2^2)$.
Notice that these skeletons are equivalent if $c_1' = 0$, $c_2' = 2\,c_1$, and $c_3' = 1$.
In addition, from the skeleton performance evaluation shown in Table~\ref{tab:skeleton_evaluation}, we verified that our method consistently attained lower or comparable error metrics compared to other SR methods. 
These results strongly support our hypothesis that our method would generate univariate skeletons that are more similar to those corresponding to the underlying equations in comparison to other SR methods.

Note that E2E produced the correct skeleton for at least one of the variables in most of the cases.
Recall that E2E generated expressions that minimize the prediction error; thus, it did not prioritize identifying the correct functional form of the variables that do not contribute substantially to the overall error. 
In addition, in some cases, E2E generated skeletons that were equivalent to the target skeletons but larger than those produced by MST.
For example, in problem E4, E2E generated the skeleton $\mathbf{\hat{e}}_{E2E}(x_1) = c_1 + c_2|c_3 + c_4\,x_1 + c_5\,x_1^2 + c_6\,x_1^3 + c_7\,x_1^4|$ for variable $x_1$, which is equivalent to the one produced by MST, $\mathbf{\hat{e}}_{MST}(x_1) = c_1 + c_2\,x_1 + c_3\,x_1^2 + c_4\,x_1^3 + c_5\,x_1^4$.
Another advantage over the other neural SR methods is that MST requires 24.2 million parameters while E2E requires 93.5 million.
NeSymReS requires 26.4 million parameters (i.e., 2.2 million more than MST), and Table~\ref{tab:skeleton_evaluation} shows that it failed to identify the correct functional form in most cases and is limited to problems with up to three variables.

It is worth pointing out that in problems E5, E6, E8, E9, and E13, some compared methods achieved low error metrics but are not comparable to the ones achieved by our method.
For example, in problem E6, E2E generated the skeleton $\mathbf{\hat{e}}_{E2E}(x_1) = c_1 + c_2\, \text{atan}(c_3 + c_4\,x_1)$ for variable $x_1$, which does not coincide with the functional form of the underlying skeleton $\mathbf{e}(x_1) = c_1\! +\! \text{tanh}(c_2\,x_1)$.
However, E2E achieved low error metrics because, during the coefficient fitting process, the GA found appropriate values for the constant that multiplies the argument of the \texttt{atan} function, stretching or compressing the curve, making it resemble the shape of \texttt{tanh} and minimizing the error.
Hence, the skeleton generated by E2E produced low error metrics and is considered to be similar to the target skeleton.
Conversely, the functional form of the skeleton generated by MST coincided with that of the target skeleton exactly (i.e., $\mathbf{\hat{e}}_{MST}(x_1) = c_1 + c_2\,\text{tanh}(c_3\, x_1)$) and thus produced significantly lower error metrics. 

We claim that our method can be regarded as a \textit{post-hoc} explainability method that generates univariate skeletons as explanations of the function approximated by a black-box regression model.
From an explainability perspective, the generation of more accurate univariate symbolic skeletons is crucial.
By producing skeletons that align more closely with the underlying data, we offer transparent insights into how each variable influences the function's response.

Finally, we applied our method to a precision agriculture problem. 
Given a field that has been split into MZs, the problem consists of determining the functional form that describes the N-response curves corresponding to all sites within an MZ. 
This is modeled as an MSSP problem and the skeleton generated for each MZ is reported in Table~\ref{tab:results_Nresponse}. 
Our results show that the generated skeletons yield lower fitting errors and, thus are more suitable when modeling the field's N-response curves.
In future work, the function fitted for each site will serve as a surrogate yield model that may aid in optimization tasks. 

One potential limitation of our approach, as well as any neural SR method, lies in its ability to generate skeletons whose complexity is bounded inherently by the expressions produced during the pre-training phase of the Multi-Set Transformer.
For instance, we would not be able to identify the skeleton $c_1 + c_2\, x_2^2/\sin(c_3\,e^{c_4\,x_2})$ as it requires eight operators, while our training set was limited to expressions with up to seven operators.
However, it is feasible to overcome this limitation through transfer learning, so that the MST model can be trained on more complex tasks, potentially enabling the recognition of such complex skeletons.
Another limitation is that we do not provide multi-variate expressions as other methods do.
This is because the objective of this work lies in discerning the functional relationship between each variable and the system's response accurately.
Future work will focus on using the generated univariate skeletons as building blocks to produce multivariate expressions that approximate the system's entire underlying function. 
For instance, mathematical expressions explaining the full behavior of real-world systems will be generated, and their performance will be evaluated by comparing observed response values with those predicted by the generated expressions.

\section{Conclusions}

Symbolic regression aims to find symbolic expressions that represent the relationships within the observed data. 
As such, SR represents a promising avenue for building explainable models. 
By seeking to uncover mathematical equations that represent the relationships between input variables and their response, the resulting equations offer transparency and clear insights into model behavior. 

Given a multivariate regression problem, the objective of the work reported in this paper is to produce symbolic skeletons that describe the relationship between each variable and the system's response.
To do so, we introduced a problem called Multi-Set Skeleton Prediction that aims to generate a symbolic skeleton expression that characterizes the functional form of multiple sets of input--response pairs.
To solve this problem, we proposed a novel Multi-Set Transformer model, which was pre-trained on a large dataset of synthetic symbolic expressions.
Then, the multivariate SR problem is tackled as a sequence of MSSP problems generated using a black-box regression model.
The generated skeletons are regarded as explanations of the black-box model’s function. 
Experimental results showed that our method consistently produced more accurate univariate skeletons in comparison to two GP-based SR methods and two neural SR methods.

Future work will focus on merging the generated univariate skeletons into a multivariate symbolic expression that approximates the underlying function of the system.
This task requires us to answer questions such as how to ensure that the skeletons generated for all variables are compatible and can be merged, or how to find the optimal order in which to merge the skeletons.

\section*{Acknowledgments}
The authors wish to thank the members of the Numerical Intelligent Systems Laboratory (NISL) for their comments throughout the development of this work.
This research was supported by the Data Intensive Farm Management project (USDA-NIFA-AFRI 2016-68004-24769 and USDA-NRCS NR213A7500013G021).
Computational efforts were performed on the Tempest HPC System, operated and supported by University Information Technology Research Cyberinfrastructure at Montana State University.

\bibliographystyle{splncs04}



\appendix

\counterwithin{figure}{section}
\counterwithin{table}{section}
\counterwithin{algorithm}{section}
\counterwithin{definition}{section}
\counterwithin{equation}{section}

\section{Set Transformer vs. Multi-Set Transformer} \label{app:SetTransformer}

Here, we provide background information about the Set Transformer, the foundational basis for the Multi-Set Transformer.
Furthermore, we enumerate the differences between the architectures and application domains of both models.

The Transformer~\cite{attention} was designed to handle sequences with fixed orderings, where each element's position is crucial to understanding the data's meaning. 
In contrast, sets are collections of elements without any inherent order, and their permutations do not alter the underlying semantics. 
This inherent permutation invariance poses a significant obstacle for traditional sequence models when processing sets~\cite{deepsets,neuralstatistician}.
In order to address the limitations of conventional approaches, Lee \textit{et al.}~\cite{settransformer} presented an attention-based neural network module called Set Transformer that is based on the transformer model. 
This method introduces modifications to the transformer architecture, enabling it to handle sets without assuming a fixed ordering of elements.

A model designed for set-input problems must meet two essential criteria to effectively handle sets: 
First, it should be capable of processing input sets of varying sizes; second, it should exhibit permutation invariance. 
The latter means the output of the function represented by the model remains the same regardless of the order in which the elements of the input set are presented. 
More formally, Zaheer \textit{et al.}~\cite{deepsets} described this type of function as permutation equivariant:

\begin{definition}
Consider an input set $\mathbf{x}=\{x_1, \dots, x_n \}$, where $x_i \in \mathscr{X}$ (i.e., the input domain is the power set $\mathcal{X} = 2^{\mathscr{X}}$). A function $u: 2^{\mathscr{X}} \rightarrow \mathcal{Y}$ acting on sets must be permutation invariant to the order of
objects in the set, i.e. for any permutation $\pi$: 
\[
u(\{x_1, \dots, x_n\}) = u(\{x_{\pi(1)}, \dots, x_{\pi(n)}\}),
\]
where $\pi \in S_n$ and $S_n$ represents the set of all permutations of indices $\{1, \dots, n \}$.
\end{definition}

The Set Transformer consists of two main parts: an encoder $\phi$ and a decoder $\psi$.
The process starts by encoding the set elements in an order-agnostic manner. 
As such, the encoding for each element should be the same regardless of its position in the set. 
Then, $\psi$ aggregates the encoded features and produces the desired output.
Let us consider an input set $\mathbf{S} = \{\mathbf{s}_1, \dots, \mathbf{s}_n \}$, where each element is $d_{in}$-dimensional ($\mathbf{S} \in \mathbb{R}^{n\times d_{in}}$). 
Therefore, the output $T$ produced by the Set Transformer, whose computed function is denoted as $g$, can be expressed as:
\begin{equation}
    T= g(\mathbf{S}) = \psi \left( \phi\left( \left\{ \mathbf{s}_1, \dots, \mathbf{s}_n \right\} \right) \right).
    \label{eq:settrans}
\end{equation}

Typically, the order-agnostic property is achieved by making $\phi$ to act on each element of a set independently (i.e., $\psi \left( \left\{ \phi(\mathbf{s}_1), \dots, \phi(\mathbf{s}_n) \right\} \right)$)~\cite{deepsets}.
However, the Set Transformer uses self-attention mechanisms to encode the entire input set simultaneously to recognize interactions among the set instances.

For example, Lee \textit{et al.}~\cite{settransformer} used self-attention blocks (SABs), which, given an input set $A$, perform self-attention between the elements of the set and produce a set of equal size.
This operation computes pairwise interactions among the elements of $A$; therefore, a stack of multiple SAB operations would encode higher-order interactions.
One drawback of the use of SABs is that the attention mechanism is performed between two identical sets with $n$ elements, which leads to a quadratic time complexity $\mathcal{O}(n^2)$.
In order to alleviate this issue, Zaheer \textit{et al.}~\cite{deepsets} introduced the induced set attention block (ISAB).
This could be interpreted as if the input set $A$ was projected into a lower dimensional space and then reconstructed to produce outputs with the desired dimensionality.
Since the attention mechanism involved in the ISAB operation is computed between a set of size $m$ and a set of size $n$, its associated time complexity is $\mathcal{O}(mn)$.
Furthermore, Lee \textit{et al.}~\cite{settransformer} proved that the SAB and ISAB blocks are universal approximators of permutation invariant functions.

It is worth noting that the Set Transformer was designed to produce fixed-size outputs.
This setting can be used for applications such as population statistic estimation (e.g., retrieving a unique value that presents the median of an input set), unique character counting (i.e., obtaining the number of unique characters in a set of images), or $k$-amortized clustering where the objective is to produce $k$ pairs of output parameters of a mixture of Gaussians (i.e., mean and covariance).  

\begin{figure}[t]
    \centering
    \includegraphics[width=\textwidth]{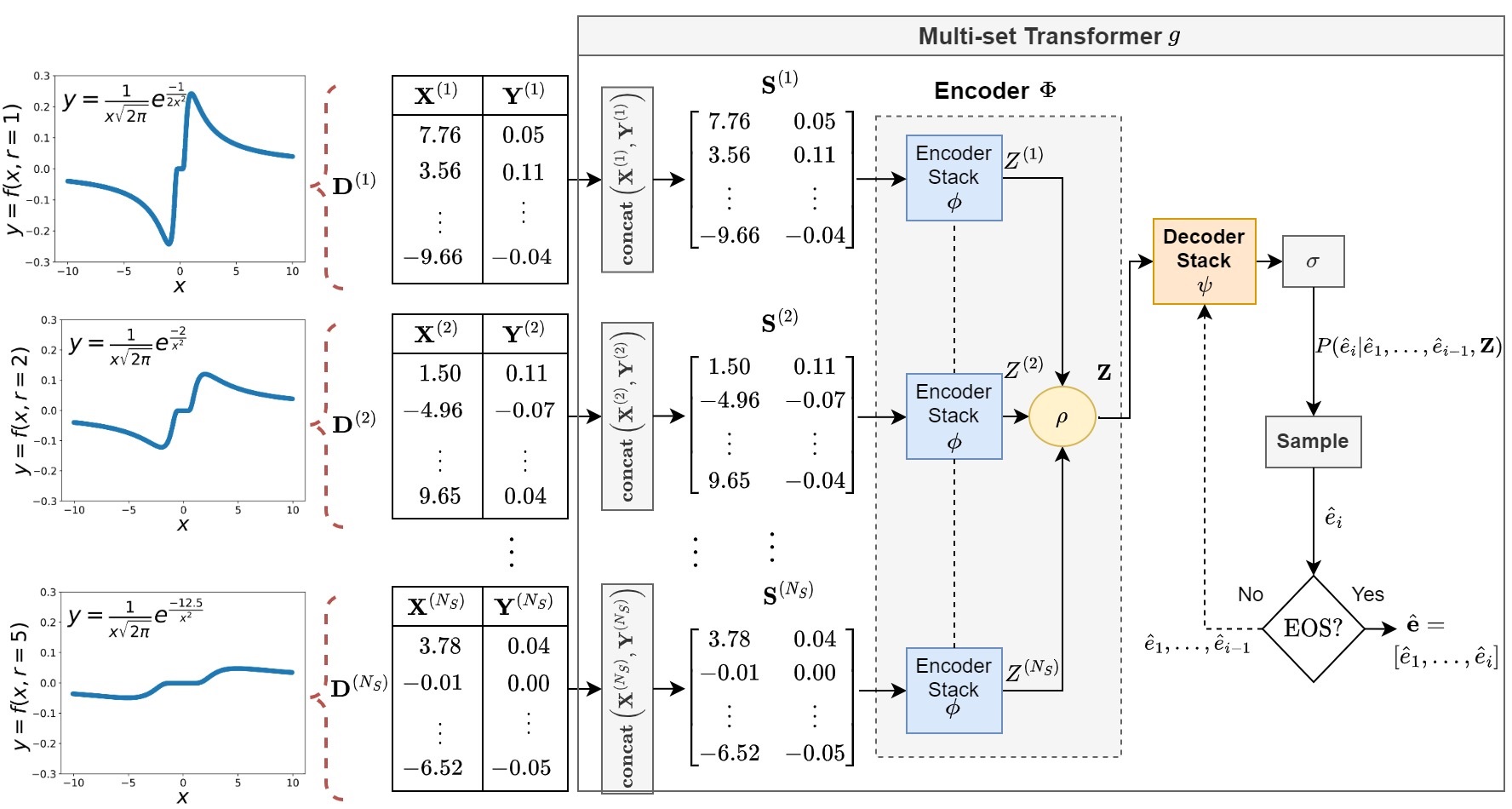}
    \caption{An example of an MSSP problem using the Multi-set Transformer. }
    \label{fig:multisetarch}
\end{figure}

The most evident limitation of the Set Transformer, in the context of the problem discussed in this paper, resides in its encoder structure, which is specifically designed to process a single input set, as shown in Eq.~\ref{eq:settrans}. 
This is important because the input of the MSSP problem is defined as a collection of $N_S$ input sets ($N_S >1$).
Hence, the encoder of our Multi-set Transformer is designed to process multiple input sets simultaneously.
The Multi-Set Transformer architecture is depicted in Fig.~\ref{fig:multisetarch}.
All input sets shown in this example were generated using the equation $y = \frac{1}{x \sqrt{2\pi}} e^{-\frac{1}{2} (\frac{r}{x})^2}$ and 
a different $r$ value was used for each set $\mathbf{D}^{(s)}$. 
Its architecture differs from the one used in the SSP method proposed by Biggio \textit{et al.}~\cite{SRthatscales}, whose encoder only processes single input sets.

Note that the Set Transformer's output is $k$-dimensional; that is, its size is fixed depending on the problem. 
Conversely, the objective of the MSSP problem is to generate a symbolic skeleton string whose length is not known \textit{a priori} and depends on each input collection.
Therefore, the decoder of the Multi-set Transformer is designed as a conditional-generative structure as it generates output sequences (i.e., the skeleton string) based on the encoded context.

\section{Data Generation} \label{app:equationgeneration}

In this section, we explain how to generate the synthetic equations and corresponding data used to train our Multi-set Transformer. 
In addition, we explain the methodology we used to avoid generating functions that lead to undefined values during the training process.

\subsection{Univariate Symbolic Skeleton Dataset Generation} \label{app:datasetgeneration}

\begin{table}[!t]
\caption{Vocabulary used to pre-train the Multi-set Transformer.}
\label{tab:vocabulary}
\centering
\setlength{\tabcolsep}{16pt}
\begin{tabular}{|c|c|c|c|c|c|}
\hline
\textbf{Token} & \textbf{Meaning} & \textbf{Token} & \textbf{Meaning} \\
\hline
\texttt{SOS} & Start of sentence &  \texttt{sin} & Sine\\
\hline
\texttt{EOS} & End of sentence &  \texttt{sinh} & Hyperbolic sine\\
\hline
\texttt{c} & Constant placeholder &  \texttt{sqrt} & Square root\\
\hline
\texttt{x} & Variable &  \texttt{tan} & Tangent\\
\hline
\texttt{abs} & Absolute value &  \texttt{tanh} & Hyperbolic tangent\\
\hline
\texttt{acos} & Arc cosine &  \texttt{-3} & Integer number\\
\hline
\texttt{add} & Sum &  \texttt{-2} & Integer number\\
\hline
\texttt{asin} & Arc sine & \texttt{-1} & Integer number \\
\hline
\texttt{atan} & Arc tangent &  \texttt{0} & Integer number\\
\hline
\texttt{cos} & Cosine &  \texttt{1} & Integer number\\
\hline
\texttt{cosh} & Hyperbolic cosine & \texttt{2} & Integer number \\
\hline
\texttt{div} & Division & \texttt{3} & Integer number \\
\hline
\texttt{exp} & Exponential & \texttt{4} & Integer number \\
\hline
\texttt{log} & Logarithmic & \texttt{5} & Integer number \\
\hline
\texttt{mul} & Multiplication & \texttt{E} & Euler's number \\
\hline
\texttt{pow} & Power & --- & ---\\
\hline
\end{tabular}
\end{table}

Table~\ref{tab:vocabulary} provides the vocabulary used in this work.
Our method differs from the generation method proposed by Biggio \textit{et al.}~\cite{SRthatscales}, which, in turn, was based on that by Lample \textit{et al.}~\cite{Lample2020Deep}.
Their method consists of constructing an expression tree with a specified number of total operators (i.e., unary and binary operators) by selecting operators and inserting them into random available places of the tree structure iteratively.
The operators are chosen based on a pre-defined set of probability weights. 
In practice, we noticed that this approach leads to the generation of several expressions that contain binary operators exclusively and, thus, can be simplified into simple expressions such as $c_1\,x$ and $c_1\,x^2 + c_2$.

As an alternative, we use a generation method that builds the expression tree recursively in a preorder fashion (i.e., the root node is created before generating its left and right subtrees recursively).
At each step, the algorithm decides whether to add a binary operator, a unary operator, or a leaf node. 
Instead of assigning specific probability weights to individual operators, we assign weights to the types of nodes to be generated, namely unary operators, binary operators, or leaf nodes, with weights of 2, 1, and 1, respectively.
In addition, rather than specifying the exact number of operators to include, we set a maximum limit on the total number of operators allowed in the expression.
This adds flexibility in cases where including additional operators may incur violations of syntaxis or pre-defined conditions.

By generating the tree in a preorder manner, we can enforce certain conditions and constraints effectively.
In particular, we can impose restrictions such as setting a maximum limit on the nesting depth of unary operators within each other.
In addition, we can forbid certain combinations of operators.
For example, we avoid embedding the operator \texttt{log} within the operator \texttt{exp}, or vice versa, since such composition could lead to direct simplification (e.g., $\texttt{log}\left( \texttt{exp} (x) \right) = x$).
We can also avoid combinations of operators that would generate extremely large values (e.g., $\texttt{exp}\left( \texttt{exp} (x) \right)$ and $\texttt{sinh} \left( \texttt{sinh} (x) \right)$). 
Table~\ref{tab:forbidden} shows the forbidden operators we considered for some specific parent operators.

\begin{table}[t]
\centering
\caption{Forbidden combinations of operators}
\label{tab:forbidden}
\begin{tabular}{|c|c|}
\hline
\textbf{Parent Node} & \textbf{Forbidden Operators} \\ \hline
\texttt{abs} & \texttt{sqrt}, \texttt{pow2}, \texttt{pow4} \\ \hline
\texttt{exp}, \texttt{tan}, \texttt{ln} & \texttt{exp}, \texttt{sinh}, \texttt{cosh}, \texttt{tanh}, \texttt{tan}, \texttt{ln}, \texttt{pow3}, \texttt{pow4}, \texttt{pow5} \\ \hline
\texttt{sinh}, \texttt{cosh}, \texttt{tanh} & \texttt{exp}, \texttt{sinh}, \texttt{cosh}, \texttt{tanh}, \texttt{tan}, \texttt{ln}, \texttt{pow2}, \texttt{pow3}, \texttt{pow4}, \texttt{pow5} \\ \hline
\texttt{pow2}, \texttt{pow3}, \texttt{pow4}, \texttt{pow5} & \texttt{pow2}, \texttt{pow3}, \texttt{pow4}, \texttt{pow5}, \texttt{exp}, \texttt{sinh}, \texttt{cosh}, \texttt{tanh} \\ \hline
\texttt{sin}, \texttt{cos}, \texttt{tan} & \texttt{sin}, \texttt{cos}, \texttt{tan} \\ \hline
\texttt{asin}, \texttt{acos}, \texttt{atan} & \texttt{asin}, \texttt{acos}, \texttt{atan} \\ \hline
\end{tabular}
\end{table}

\begin{figure}[!ht]
    \centering
    \includegraphics[width=\textwidth]{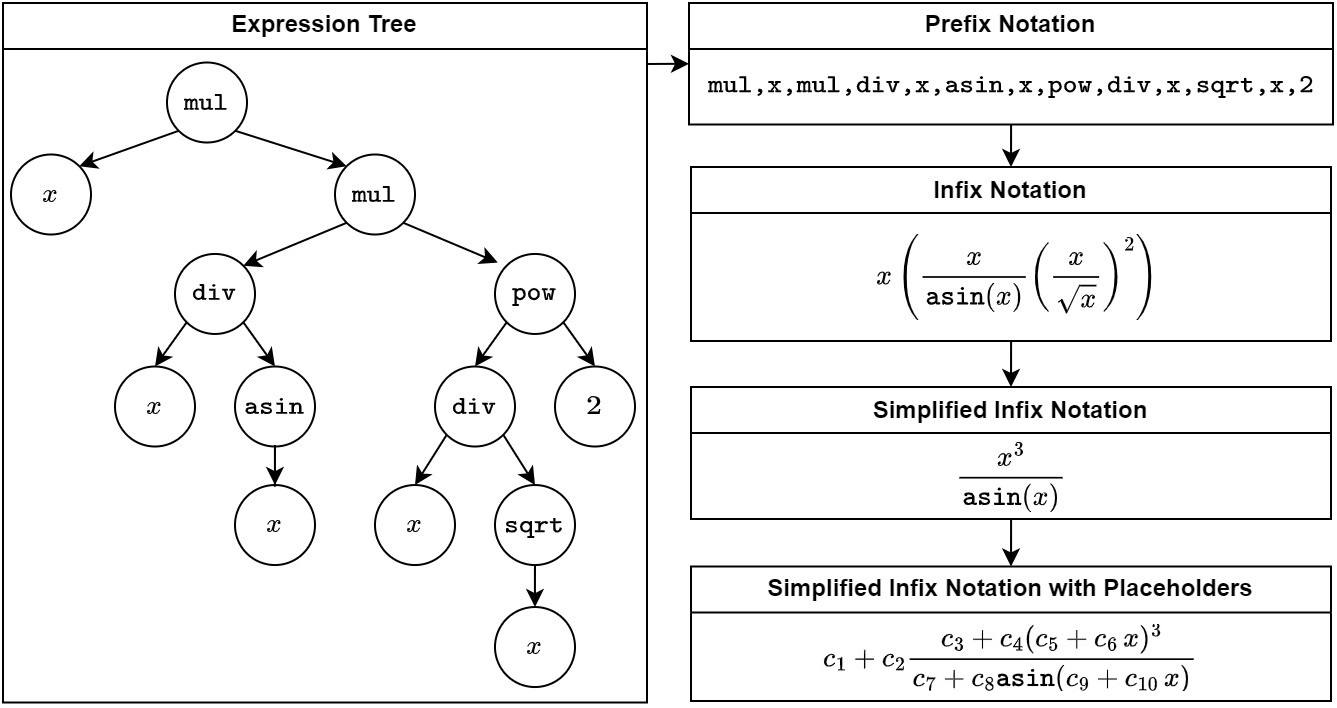}
    \caption{Example of a randomly generated expression.}
    \label{fig:eqgeneration}
\end{figure}

Each generated tree is traversed to derive a mathematical expression in prefix notation.
The expression is then transformed from prefix to infix notation, which is simplified using the SymPy library\footnote{\url{https://www.sympy.org/}}.
The simplified expression is transformed again into an expression tree.
Then, each non-numerical node of the tree is multiplied by a unique placeholder and then added to another unique placeholder.
Some exceptions apply; for instance, the arguments of the $\texttt{exp}$, $\texttt{sinh}$, $\texttt{cosh}$, and $\texttt{tanh}$ operators are not affected by additive placeholders.
This is because adding a constant to their arguments would lead to direct simplification (e.g., $c_1 e^{c_2\,x + c_3} = (c_1 e^{c_2}) e^{c_3 \,x}$).
Fig.~\ref{fig:eqgeneration} illustrates the random expression generation process.
Note that previous approaches also proposed to include unique placeholders in the generated expressions; however, they included only one placeholder at a time and in some of the nodes.
In contrast, our approach allows for the generation of more general expressions.

\subsection{Avoiding Invalid Operations} \label{app:avoidNaNs}

In this section, we present a detailed description of the function $\texttt{avoidNaNs}(\mathbf{x}, f)$, initially introduced in Algorithm 1 of the main paper.
This function serves the purpose of modifying certain coefficients within the function $f$ or generating supplementary support values within the vector $\mathbf{x}$ to avoid numerical inconsistencies.
We classify the operators that may generate undefined values as follows:

\begin{itemize}
    \item Single-bounded operators: These operators have bounded numerical arguments due to the mathematical constraints imposed on their domains. We consider the following single-bounded operators:
\begin{itemize}
    \item Logarithm: The \texttt{log} operator is bounded on its left side as it cannot receive an input lower than or equal to 0:
\[\text{Domain}\left( \texttt{log}(x) \right) = \{ x \in \mathbb{R} \; | \; x>0 \}.\]
    \item  Square root:  The \texttt{sqrt} operator  is also bounded on its left side to avoid generating complex numbers:
\[\text{Domain}\left( \texttt{sqrt}(x) \right) = \{ x \in \mathbb{R} \; | \; x \geq0 \}.\]
\item Exponential: We decided to bound the \texttt{exp} operator on its right side to avoid generating extremely large values. This also applies to the \texttt{sinh}, \texttt{cosh}, and \texttt{tanh} operators: 
\[\text{Domain}\left( \texttt{exp}(x) \right) = \{ x \in \mathbb{R} \; | \; x \leq 7 \}.\]
\end{itemize}
    \item Double-bounded operators: Unlike the single-bounded operators, the numerical arguments of these operators are bounded on their left and right sides. We consider the following double-bounded operators:
    
\begin{itemize}
    \item Arcsine: The \texttt{asin} operator takes a value between -1 and 1 as its input and returns the angle whose sine is equal to that value:
\[\text{Domain}\left( \texttt{asin}(x) \right) = \{ x \in \mathbb{R} \; | \; -1 \leq x \leq 1 \}.\]
    \item  Arccosine:  Like arcsine, the \texttt{acos} also takes a value between -1 and 1 as its input operator:
\[\text{Domain}\left( \texttt{acos}(x) \right) = \{ x \in \mathbb{R} \; | \; -1 \leq x \leq 1 \}.\]
\end{itemize}

    \item Operators with singularities: Operators like tangent or division can exhibit singularities at specific input values. For instance, the tangent function becomes undefined when its input equals an odd multiple of  $\pi/2$, resulting in an asymptotic behavior where the function approaches infinity.

\end{itemize}

The method proposed to address the NaN (``not a number") values arising from the aforementioned functions, which we refer to as ``special functions", is shown in Algorithm~\ref{alg:avoidNaNs}.
Here, $f.\texttt{args}$ returns a vector containing the numerical arguments of function $f$. 
If $f$ is considered as an expression tree, $f.\texttt{func}$ returns the name of the operator located at the top of the tree.
The algorithm analyzes each argument of $f$ separately (Line 3).
A given argument of $f$, denoted as $arg$, can be considered as a sub-expression tree.
Hence, the function $\texttt{containSpecialF}(arg)$ (Line 4) traverses the sub-expression tree and returns a true value if any special function is found in it.

If the current sub-expression tree contains a special function, we check if the top operator of the subtree is special using function $\texttt{isSpecialF}(arg.\texttt{func})$ (Line 5). 
If not, we move down to a deeper level of the subtree using recursion (Line 6).
Otherwise, the arguments of the sub-expression $arg$ may need to be modified to avoid undefined values.
Before doing so, in Line 8, we verify if there is another special function contained inside the current subtree, in which case we explore a deeper level of the subtree using recursion.
Note that we apply the $\texttt{containSpecialF}(arg.\texttt{args}[0])$ function given that $arg$ is guaranteed to represent a special function; i.e., it is a unary function with a single argument.

\begin{figure}[!t]
    \centering
    \includegraphics[width=\textwidth]{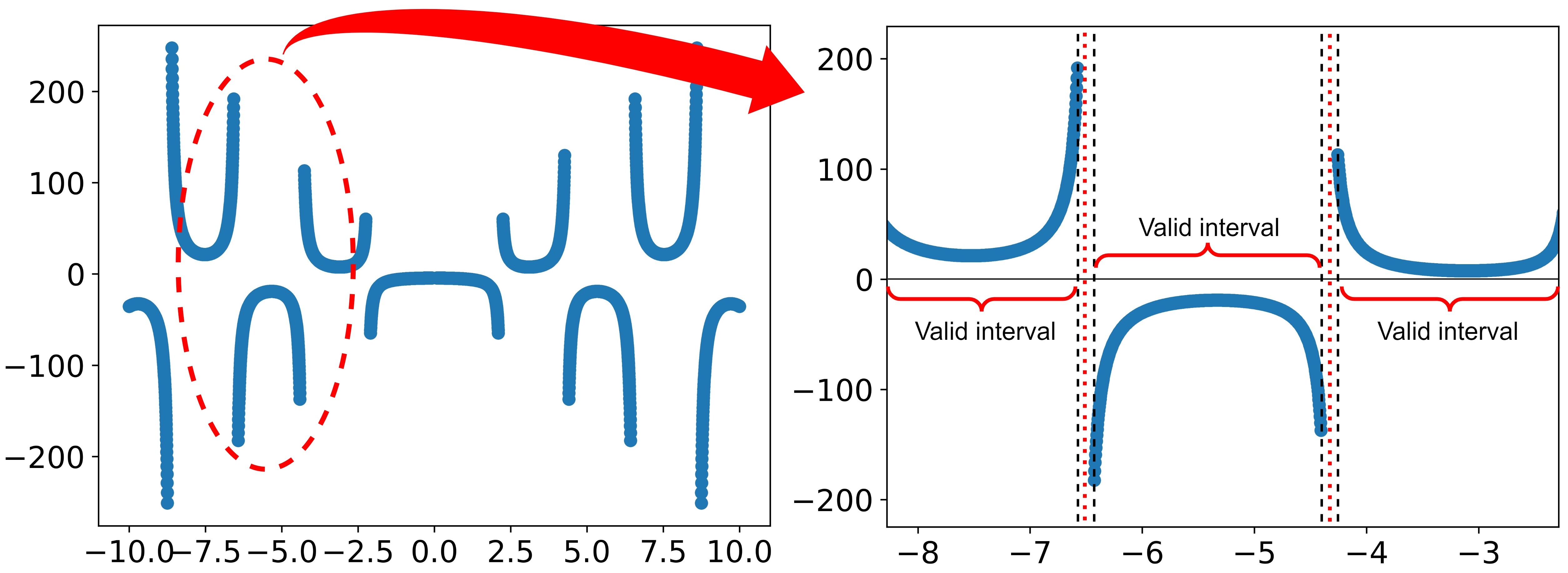}
    \caption{Generation data for $f(x) = \frac{-3.12 x}{\texttt{sin}(1.45 x)} - 2.2$. (Left) Generated data on the entire domain. (Right) Detailed view of how singularities are avoided.}
    \label{fig:singularity}
\end{figure}

The function $arg$ is evaluated on the input values $\mathbf{x}$ and the obtained values are stored in the vector $vals$ (Line 11).
The function $\texttt{containNaNs}(vals)$ returns a true value in the event that one or more undefined values are found within the vector  $vals$.
If an undefined value is found, a modification of the inner argument, $innArg$, or the input vector $\mathbf{x}$ is needed.
To do so, we first evaluate the function represented by $innArg$ on the input values $\mathbf{x}$ and store the outcomes in the variable $innVals$ (Line 13).
In the case that the function $arg$ is single-bounded, the domain of function $innArg$ is modified accordingly using $\texttt{modifySBounded}(arg, vals, innArg, innVals)$ (Line 15) by adding a horizontal offset.
Likewise, if  $arg$ is double-bounded, the domain of $innArg$ is modified accordingly using $\texttt{modifyDBounded}(arg, innArg, innVals)$ (Line 17) by re-scaling $innArg$.
In the case that $arg$ represents a function with singularities, a new input vector $\mathbf{x}$ with resampled values is obtained using the function $\texttt{avoidSingularities}(\mathbf{x}, arg, Xsing)$ (Line 19), where $Xsing$ is a variable that will store all positions at which the function produces undefined values.
Fig.~\ref{fig:singularity} depicts an example of how data are generated using the underlying function $f(x) = \frac{-3.12 x}{\texttt{sin}(1.45 x)} - 2.2$.
The figure at the bottom shows in detail that the $\texttt{avoidSingularities}$ function generates valid intervals of values for $\mathbf{x}$ by avoiding getting too close to the singular points (red dotted line).
Furthermore, the inner argument of function $arg$ is then replaced with the modified function $innArg$ (Line 21).
Finally, the arguments of the original function $f$ are replaced with the modified arguments stored in the list $newArgs$ (Line 22). 

\begin{algorithm} [!t]
\scriptsize
\begin{algorithmic}[1]
\Function{avoidNaNs}{$\mathbf{x}, f, Xsing = [\,]$}
    \State $args, newArgs \leftarrow f.\texttt{args}, []$
    \For { each $arg \in args$}
            \If {$\texttt{containSpecialF}(arg)$} \label{alg:contains}            
                    \If {$! \texttt{isSpecialF}(arg.\texttt{func})$}  \Comment{If the top operator is special}
                            \State $\mathbf{x}, arg, Xsing \leftarrow \texttt{avoidNaNs}(\mathbf{x}, arg, Xsing)$
                    \Else 
                            \If {$\texttt{containSpecialF}(arg.\texttt{args}[0])$} \Comment{If there's a special function inside}
                                    \State $\mathbf{x}, arg, Xsing  \leftarrow \texttt{avoidNaNs}(\mathbf{x}, arg, Xsing )$                         
                            \EndIf
                            \State $innArg \leftarrow arg.\texttt{args}[0]$
                            \State $vals \leftarrow arg(\mathbf{x})$
                            \If{$\texttt{containNaNs}(vals)$}
                                    \State $innVals \leftarrow innArg(\mathbf{x})$
                                    \If{$\texttt{isSingleBounded}(arg)$}
                                            \State $innArg \leftarrow \texttt{modifySBounded}(arg, vals, innArg, innVals)$
                                    \ElsIf{$\texttt{isDoubleBounded}(arg)$}
                                            \State $innArg \leftarrow \texttt{modifyDBounded}(arg, innArg, innVals)$
                                    \Else  \; \Comment{Operations with singularities}
                                            \State $\mathbf{x}, sing \leftarrow \texttt{avoidSingularities}(\texttt{length}(\mathbf{x}), arg, Xsing)$
                                    \EndIf
                            \EndIf
                            \State $arg.\texttt{args}[0] \leftarrow innArg$ \Comment{Update function}
                    \EndIf
            \EndIf
            \State $newArgs.\texttt{append}(arg)$
    \EndFor
    \State $f.\texttt{args} \leftarrow newArgs$
    \State \Return $\mathbf{x}, arg, sing$
\EndFunction
\end{algorithmic}
\caption{Avoiding Invalid Operations}
\label{alg:avoidNaNs}
\end{algorithm}

\section{Comparison of Predicted Symbolic Skeletons} \label{app:results}

In this section, we report the symbolic skeletons predicted by two GP-based SR methods, PySR~\cite{pysr2} and TaylorGP~\cite{TaylorGP}, and three neural SR methods, NeSymRes~\cite{SRthatscales}, E2E~\cite{SRtransformer}, and our Multi-Set Transformer.

Tables~\ref{tab:results_skeletons} and \ref{tab:results_skeletons2} report the skeletons obtained by all compared methods.
It is important to note that these experiments are designed to validate the functional form obtained between each system's variable and the system's response.
That is, instead of evaluating the expressions generated by the compared methods directly, we first extract their skeletons with respect to each system's variable.
For example, NeSymRes produces the estimated function $\hat{f}(x_1, x_2) = 0.592\,x_1\,x_2 + \cos(0.014\,(-0.995\,x_1 + x_2 - 0.08)^2)$ for problem E1.
From this, the skeletons with respect to variables $x_1$ and $x_2$ are $\hat{e}(x_1)=c_1\, x_1 + \cos(c_2 (c_3 + c_4\, x1)^2)$ and $\hat{e}(x_2)= c_1\, x_2 + \cos(c_2 (c_3 + x_2)^2)$, respectively.

\begin{landscape}
\begin{table}[t]
    \caption{Comparison of skeleton prediction results (E1--E9)}
    \label{tab:results_skeletons}
\centering
\Large
\resizebox{19cm}{!}{%
\def\arraystretch{1.5}
\begin{tabular}{|c|c|c|c|c|c|c|c|}
\hline
\textbf{Prob.} & \diagbox[width=25mm]{\textbf{Var.}}{\textbf{Met.}} & \textbf{PySR} & \textbf{TaylorGP} & \textbf{NeSymReS} & \textbf{E2E} & \textbf{MST} & \textbf{Target} $\mathbf{e}(x)$ \\ \Xhline{4\arrayrulewidth}
\multirow{2}{*}{\textbf{E1}} & $x_1$ & $c_1\, x_1$& $c_1\, x_1$& $c_1\, x_1 + \cos(c_2 (c_3 + c_4\, x_1)^2)$& $c_1 + c_2\, x_1 + c_3 \sin((c_4 + c_5 \, x_1)^2)$& $c_1 + c_2\, x_1 + c_3 \sin(c_4 + c_5\, x1)$& $c_1\,x_1 + c_2\, \sin(c_3(c_4 + x_1))$\\ \cline{2-8} 
 & $x_2$ & $c_1\, x_2$& $c_1\, x_2$& $c_1\, x_2 + \cos(c_2 (c_3 + x_2)^2)$& $c_1 + c_2(c_3 + c_4\, x_2)$& $c_1 + c_2\, x_2 + c_3 \sin(c_4 + c_5\, x_2)$& $c_1\, x_2 + c_2\, \sin(c_3(c_4 + x_2))$\\ \Xhline{4\arrayrulewidth}
\multirow{3}{*}{\textbf{E2}} & $x_1$ & $c_1 +\left|c_2 + |c_3 + x_1|\right|$ & $c_1 + c_2 \,x_1$& $c_1 + c_2\,x_1$ & $c_1 + c_2\, x_1 + c_3(c_4 + c_5\, x_1)^2$ & $c_1 + c_2(c_3 + c_4\, x_1)^2$ & $c_1 + (c_2 + c_3\,x_1)^2$ \\ \cline{2-8} 
 & $x_2$ & $c_1$ & $c_1$& $c_1 + e^{e^{c_2\,x_2}}$ & $c_1 + c_2\,(c_3 + c_4\, x_2)$ & $c_1\sqrt{c_2\,x_2 + c_3} + c_4$ & $c_1\sqrt{x_2 + c_2} + c_3$ \\ \cline{2-8} 
 & $x_3$ & $c_1 + c_2\,x_3$ & $c_1 + c_2 \, x_3$ & $c_1 + c_2 \, x_3$ & $c_1 + c_2 \, x_3 + c_3 (c_4 + c_5 \cos(c_6 + c_7 \, x_3))$ & $c_1 + c_2 \sin(c_3\, x_3 + c_4)$& $c_1 + c_2\sin(c_3\,x_3)$ \\ \Xhline{4\arrayrulewidth}
\multirow{2}{*}{\textbf{E3}} & $x_1$ &  $c_1 \,e^{x_1}\, |\sinh(c_2\, x_1)|$& $c_1 \, e^{c_2 \, x_1}$&  $c_1 \, e^{c_2 \, x_1}$&  $c_1 + c_2 e^{c_3\, x_1}$&  $c_1 + c_2 e^{c_3\, x_1}$&  $c_1 + c_2\,e^{c_3\,x_1}$\\ \cline{2-8} 
 & $x_2$ &  $c_1$ & $c_1$ &  $c_1\, \cos(c_2\, x_2)$&  $c_1 + c_2\, \cos(c_3 + c_4\, x_2)$&  $c_1 + c_2\, \cos(c_3 + c_4\, x_2)$&  $c_1 + c_2\, \cos(c_3\, x_2)$\\ \Xhline{4\arrayrulewidth}
\multirow{4}{*}{\textbf{E4}} & $x_1$ &  $c_1 + c_2\, x_1^2$
& $c_1\, x_1^2$&  ---
&  $c_1 + c_2 |c_3 + c_4\,x_1 + c_5\,x_1^2 + c_6\,x_1^3 + c_7\,x_1^4|$&  $c_1 (c_2\, x1 + c_3)^4 + c_4$&  $c_1 + c_2\,x_1 + c_3\, x_1^2 + c_4\, x_1^4$\\ \cline{2-8} 
 & $x_2$ &  $c_1 + c_2\, x_2$& $c_1$ &  ---
&  $c_1 + c_2|c_3 + c_4\, x_2 + c_5\, x_2^2|$&  $c_1 (c_2\, x2 + c_3)^2 + c_4$&  $c_1 + c_1\, x_2 + c_3\, x_2^2$\\ \cline{2-8} 
 & $x_3$ & $c_1 + c_2\, x_3^2$ & $c_1$ & \multicolumn{1}{c|}{---
} & \multicolumn{1}{l|}{$c_1 + c_2 |c_3 + c_4\,x_3 + c_5\,x_3^2 + c_6\,x_3^3 + c_7\,x_3^4|$} & $c_1 (c_2\, x3 + c_3)^4 + c_4$ & $c_1 + c_2\, x_3 + c_3\, x_4^2 + c_4\, x_3^4$ \\ \cline{2-8} 
 & $x_4$ & $c_1 + c_2\, x_4$ & $c_1$ & \multicolumn{1}{c|}{---
} & $c_1 + c_2|c_3 + c_4\, x_4 + c_5\, x_4^2|$ & $c_1 (c_2\, x4 + c_3)^2 + c_4$ & $c_1 + c_2\, x_4 + c_3\, x_4^2$ \\ \Xhline{4\arrayrulewidth}
\multirow{4}{*}{\textbf{E5}} & $x_1$ &  $c_1$ & $c_1$ &  ---
&  $c_1 + c_2 \cos(c_3 + c_4\,x_1)$&  $c_1 \sin(c_2 \, x1 + c_3) + c_4$&  $c_1 + \sin(c_2 + c_3\, x_1)$\\ \cline{2-8} 
 & $x_2$ &  $c_1$ & $c_1$ &  ---
&  $c_1 + c_2 \, x_2$&  $c_1 \sin(c_2 \, x2 + c_3) + c_4$&  $c_1 + \sin(c_2 + c_3\,x_2)$\\ \cline{2-8} 
 & $x_3$ & \multicolumn{1}{c|}{$c_1$ } & $c_1$ & \multicolumn{1}{c|}{---
} & $c_1 + c_2 \, x_3 + c_3 \, x_3^2 + c_4\, x_3^3$ & $c_1 \sin(c_2 \, x3 + c_3) + c_4$ & $c_1 + \sin(c_2 + c_3 \, x_3)$ \\ \cline{2-8} 
 & $x_4$ & \multicolumn{1}{c|}{$e^{c_1 \, x_4}$} & $c_1\,e^{x_4}\,e^{\sin(c_2\,x_4)}$& \multicolumn{1}{c|}{---} & $c_1 + c_2 \, e^{c_3 \, x_4}$ & $c_1 e^{c_2\, x4} + c_3$ & $c_1 + e^{c_2\, x_4}$ \\ \Xhline{4\arrayrulewidth}
 \multirow{3}{*}{\textbf{E6}} & $x_1$ &  $c_1$ &  $c_1$ & $c_1 + c_2\, x_1$&  $c_1 + c_2\, \text{atan}(c_3 + c_4\,x_1)$ &  $c_1 + c_2\, 
\text{tanh}(c_3\, x_1)$  & $c_1 + \text{tanh}(c_2\, x_1)$ \\ \cline{2-8} 
 & $x_2$ & $c_1$  & $c_1$ & $c_1 + x_2\,\sin(c_2/(c_3 + x_2))$  &  $c_1 + c_2\,x_2 + c_3\,x_2\,\cos(c_4 + c_5\,x_2)$  & $c_1 + c_2\,|x_2|$ &  $c_1 + c_2\,|x_2|$\\   \cline{2-8} 
 & $x_3$ & $\text{tanh}(\exp(x_3))$  & $c_1\,\sin(c_2\,x_3^2)/(c_3\,\sqrt{x_3} + \sin(\sqrt{x_3}))$ & $c_1$ & $c_1$  & $c_1\, \cos(c_2\,(c_3 + x_3)^2) + c_4$ &  $c_1 + c_2\,(\cos(c_3\,x_3^2))$\\  \Xhline{4\arrayrulewidth}
  \multirow{2}{*}{\textbf{E7}} & $x_1$ & $c_1 + x_1^2$  & $c_1$ & $c_1 + c_2\,x_1$  & $c_1 + c_2\, \sin(c_3 + c_4\,x_1)^2 + c_5\, \sin(c_6 + c_7\, x_1)$  &  
 $c_1 + c_2 /(c_3 + \sin(c_4\,x_1 + c_5))$ & $c_1/(c_2 + \sin(c_3\,x_1))$ \\ \cline{2-8} 
 & $x_2$ & $c_1/\text{sinh}(\text{sinh}(\text{tanh}(e^{\text{sinh}(\sin(c_2\,x_2))})))$ & $c_1\,x_2^2 + \sqrt{|x_2|}$ &  $(c_1 + x_2^2)/(c_2 + \cos(c_3\,(c_4 + c_5\,x_2)^3))$ & $c_1\,(c_2 + c_3\,x_2)^2$  & $c_1 + c_2\,x_2^2$ &  $c_1 + c_2\,x_2^2$\\  \Xhline{4\arrayrulewidth}
 
  \multirow{2}{*}{\textbf{E8}} & $x_1$ & $c_1 + \text{tanh}(c_2 + \text{cosh}(x_1))$  & $c_1$ & $e^{c_1\,\cos(c_2/x_1)}$  & $c_1 + c_2\,e{c_3\,|c_4 + c_5\,x_1|}$  & $c_1 + c_2/(c_3\, x_1^4 + c_4)$  & $c_1 + c_2\,x_1^4/(c_3 + c_4\,x_1^4)$\\ \cline{2-8} 
 & $x_2$ & $c_1 + \cos(\tan(\text{tanh}(c_2/x_2^2)))$  &  $c_1$ & $c_1 + \cos(1/x_2)$ &  $c_1 + c_2\, e^{c_3\,|c_4 + x_2|}$  &  $c_1 + c_2/(c_3\, x_1^4 + c_4)$  &  $c_1 + c_2\,x_2^4/(c_3 + c_4\,x_2^4)$\\  \Xhline{4\arrayrulewidth}
 
  \multirow{2}{*}{\textbf{E9}} & $x_1$ & $\log(c_1/(c_2 + c_3\,x_1^2))$ & $c_1 + \log(c_2/|x_1|)$ & $\log(c / |x_1|)$  & $c_1 + c_2\,\log(c_3 + c_4\,x_1 + c_5\,x_1^2)$  & $c_1 + \log(c_2\, x_1^2 + c_3)$  & $c_1 + c_2\, \log(c_3 + c_4\,x_1^2)$\\ \cline{2-8} 
 & $x_2$ & $\log(c_1 + c_2\,x_2)$ & $c_1\exp(c_2\,x_2)$ & $\log(c \, |x_2|)$ & $c_1 + c_2\,\log(c_3 + c_4/(c_5 + c_6\,x_2 + c_7\,x_2^2))$ & $c_1 + \log(c_2 + c_3\,x_2)$  &  $c_1 + \log(c_2 + c_3\,x_2)$\\  \hline
\end{tabular}%
}
\end{table}
\end{landscape}

\begin{table}[t]
    \caption{Comparison of skeleton prediction results (E10--E13)}
    \label{tab:results_skeletons2}
\centering
\Large
\resizebox{\textwidth}{!}{%
\def\arraystretch{1.5}
\begin{tabular}{|c|c|c|c|c|c|c|c|}
\hline
\textbf{Prob.} & \textbf{Var.} & \textbf{PySR} & \textbf{TaylorGP} & \textbf{NeSymReS} & \textbf{E2E} & \textbf{MST} & \textbf{Target} $\mathbf{e}(x)$ \\ \Xhline{4\arrayrulewidth}
\multirow{2}{*}{\textbf{E10}} & $x_1$ & $\sin(c_1\,x_1)$ & $c_1\,x_1$ & $\sin(c_1\,x_1)$ & $c_1 + c_2\,\sin(c_3 + c_4\,x_1)$ & $c_1 + c_2\,\sin(c_3\,x_1)$ & $\sin(c_1\,x_1)$\\ \cline{2-8} 
 & $x_2$ & $\sin(c_1\,e^{x_2})$ & $c_1\,e^{c_2\,\sqrt{|x_2|}}$ & $\sin(c_1\,e^{x_2})$ & $c_1 + c_2\,\sin(c_3 + c_4\,e^{c_5\,x_2})$ & $c_1 + \sin(c_2\,e^{c_3\, x_2})$ & $\sin(c_1\,e^{x_2})$\\ \Xhline{4\arrayrulewidth}
  \multirow{2}{*}{\textbf{E11}} & $x_1$ & $c_1\,x_1$ & $c_1\,x_1$ & $c_1\,x_1$ & $c_1\,x_1$ & $c_1 + c_2\,x_1$  & $c_1\,x_1$\\ \cline{2-8} 
 & $x_2$ & $c_1\,\log(x_2^4)$ & $c_1\,\log(|x_2|)$ & $c_1\,\log(x_2^4)$ & $c_1 + c_2\,\log(c_3 + c_4\,|c_5 + c_6\,x_2|)$ & $c_1 + c_2\,\log(c_3\,x_2^2)$ &  $c_1\,\log(x_2^4)$\\  \Xhline{4\arrayrulewidth}
  \multirow{2}{*}{\textbf{E12}} & $x_1$ & $c_1 + \sin(x_1)\,|x_1|$ & $c_1 + c_2\,x_1$ & $c_1 + c_2\,x_1$ & $c_1 + c_2\,x_1$ & $c_1 + c_2\,x_1$ & $c_1 + c_2\,x_1$\\ \cline{2-8} 
 & $x_2$ & $c_1 + c_2\, \sin(c_3/x_2)$ & $\begin{array}{ccc}c_1\, \sin(c_2/x_2) +\\ \sqrt{x_2}\,\sin(c_3/x_2)\end{array}$ & $\begin{array}{ccc}c_1\,\sin(1/x_2) +\\ x_2\,\sin(1/x_2)\end{array}$ & $c_1\,\sin(c_2/(c_3 + c_4\,x_2))$  & $c_1 + c_2\, \sin(c_3/x_2)$  & $\begin{array}{ccc}c_1 +\\ c_2\, \sin(1/x_2)\end{array}$ \\  \Xhline{4\arrayrulewidth}
  \multirow{2}{*}{\textbf{E13}} & $x_1$ & $c_1\,\sqrt{x_1}$ & $c_1 + c_2\,\sqrt{e^{\sqrt{x_1}}}$ & $c_1 + c_2\,x_1$ & $c_1 + c_2\,\log(c_3 + c_4\,x_1)$ & $c_1 + c_2\,\sqrt{c_3 + x_1}$ & $c_1\,\sqrt{x_1}$\\ \cline{2-8} 
 & $x_2$ & $c_1\,\log(x_2^2)$ & $\begin{array}{ccc}c_1 + \log(|x_2|) +\\ c_2\,\log(|x_2|)\end{array}$& $c_1\,\log(x_2^2)$ & $c_1 + c_2/(c_3 + c_4\,|c_5 + c_6\,x_2|)$ & $c_1 + c_2\,\log(c_3\,(c_4 + x_2)^2)$ &  $c_1\,\log(x_2^2)$\\  \hline
\end{tabular}%
}
\end{table}

\end{document}